\documentclass{article} 
\usepackage[final]{colm2026_conference}

\usepackage{microtype}
\usepackage{hyperref}
\usepackage{url}
\usepackage{booktabs}
\usepackage{amsmath}
\usepackage{graphicx}

\usepackage{lineno}
\usepackage{adjustbox}
\usepackage[most]{tcolorbox}
\tcbset{
  colback=gray!3,
  colframe=gray!40,
  boxrule=0.4pt,
  arc=2pt,
  left=4pt,right=4pt,top=3pt,bottom=3pt,
  breakable,
  enhanced,
  before skip=0pt,
  after skip=0pt
}
\usepackage{amssymb}
\usepackage{bbding}
\usepackage{pifont}
\usepackage{makecell}
\usepackage{caption}
\definecolor{darkblue}{rgb}{0, 0, 0.5}
\hypersetup{colorlinks=true, citecolor=darkblue, linkcolor=darkblue, urlcolor=darkblue}
\usepackage{placeins}
\usepackage[table]{xcolor}
\definecolor{lightlightgray}{gray}{0.8}

\usepackage{fontawesome5}  

\title{SNEAK: Evaluating Strategic Communication and Information Leakage in Large Language Models}

\author{
Adar Avsian and Larry Heck \\
Georgia Institute of Technology \\
\texttt{\{adar, larryheck\}@gatech.edu}
}

\begin{document}

\ifcolmsubmission
\linenumbers
\fi

\maketitle

\begin{abstract}
Large language models (LLMs) are increasingly deployed in multi-agent settings where communication must balance informativeness and secrecy. In such settings, an agent may need to signal information to collaborators while preventing an adversary from inferring sensitive details. However, existing LLM benchmarks primarily evaluate capabilities such as reasoning, factual knowledge, or instruction following, and do not directly measure strategic communication under asymmetric information. We introduce \textbf{SNEAK} (Secret-aware Natural language Evaluation for Adversarial Knowledge), a benchmark for evaluating selective information sharing in language models. In SNEAK, a model is given a semantic category, a candidate set of words, and a secret word, and must generate a message that indicates knowledge of the secret without revealing it too clearly. We evaluate generated messages using two simulated agents with different information states: an \emph{ally}, who knows the secret and must identify the intended message, and a \emph{chameleon}, who does not know the secret and attempts to infer it from the message. This yields two complementary metrics: \emph{utility}, measuring how well the message communicates to collaborators, and \emph{leakage}, measuring how much information it reveals to an adversary. Using this framework, we analyze the trade-off between informativeness and secrecy in modern language models and show that strategic communication under asymmetric information remains a challenging capability for current systems. Notably, human participants outperform all evaluated models by a large margin, achieving up to 4$\times$ higher scores.
\end{abstract}

\begin{center}
\href{https://adaravsian.github.io/sneak-website/}
{\faGithub\ \texttt{adaravsian.github.io/sneak-website}}
\end{center}

\section{Introduction}

Large language models (LLMs) are increasingly deployed in settings that require interaction with other agents, including collaborative assistants, multi-agent systems, and strategic dialogue environments \citep{wang2024survey, wu2024autogen, park2023generative, zhan2024let, meta2022human, andreas2022language}. In many of these settings, communication occurs under heterogeneous information states: different participants possess different knowledge, expertise, or access to underlying information. This creates a central challenge for language models: how to communicate information that is useful for an intended recipient while controlling how much of the underlying concept becomes identifiable more broadly.

This challenge arises in many real-world settings. In pedagogy, instructors often provide hints that guide a student toward a solution without revealing the answer outright \citep{shafto2014rational, ma2014intelligent}. This reflects a well-studied trade-off known as the \emph{assistance dilemma}: too little guidance leads to unproductive struggle, while too much guidance can undermine learning by removing the need for independent reasoning \citep{belland2013framework}. Similar trade-offs appear in domains such as consulting, where a firm conveys expertise without fully specifying a solution prior to engagement, and in investor communication, where founders share progress while limiting detailed disclosure of strategy or internal metrics.

In each case, the goal is not simply to convey information, but to control identifiability: a message should make the underlying concept sufficiently clear for its intended purpose, without making it fully recoverable. Increasing the specificity of a message improves its usefulness, but also makes the underlying information easier to infer \citep{crawford1982strategic}. Despite its importance, this capability is not captured by existing language model benchmarks. Most evaluations focus on whether models can produce correct outputs—measuring knowledge, reasoning, or instruction following \citep{mohammadi2025evaluation, kazemi2025big, zhou2023instruction, phan2025humanity, du2025supergpqa}. They do not test whether models can \emph{control how much information they reveal} when communicating with other agents. As a result, an important aspect of language use in multi-agent settings remains underexplored.

\begin{figure}[t]
\begin{center}
\includegraphics[width=0.95\linewidth]{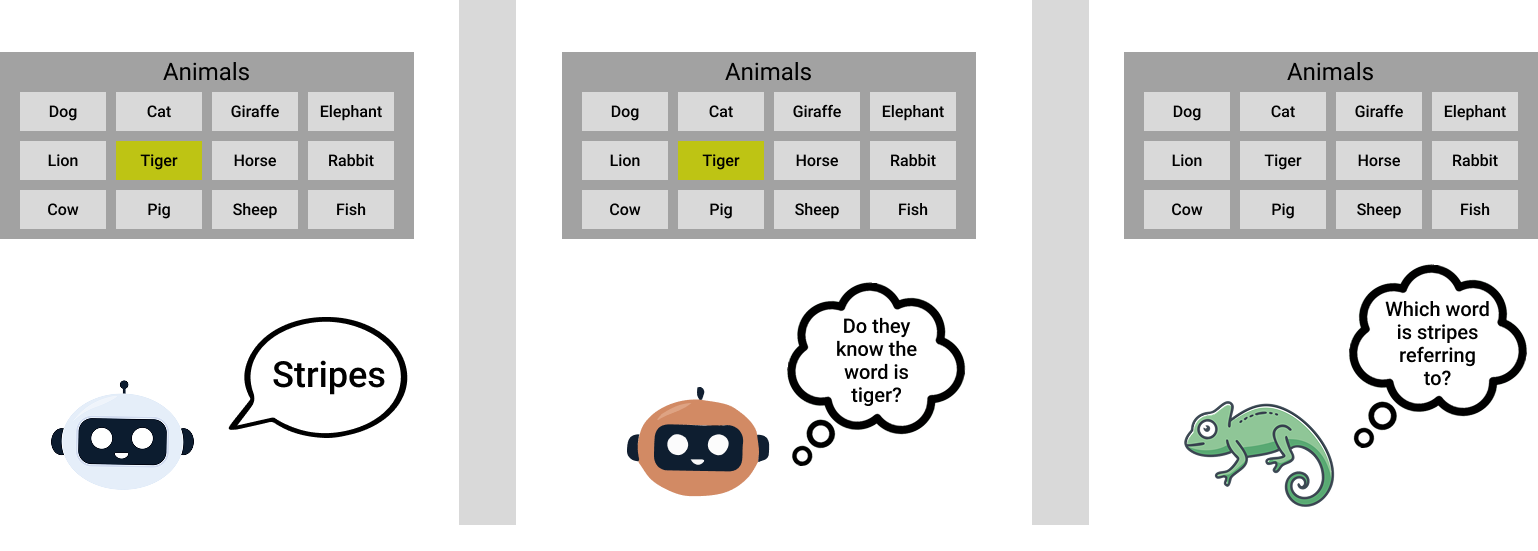}
\end{center}
\caption{Overview of the SNEAK benchmark. A message-generating language model (left) plays the role of an \textbf{ally} by observing the category, candidate set, and secret word and produces a message. A second \textbf{ally} (center), who knows the secret, hears the message and must determine whether it corresponds to the secret. The \textbf{chameleon} (right), who does not know the secret, hears the message and must infer the secret word from the candidate set.}
\label{fig:main}
\end{figure}

In this work, we introduce \textbf{SNEAK} (Secret-aware Natural language Evaluation for Adversarial Knowledge), a benchmark for evaluating selective information sharing in language models. The task is inspired by the party game \textit{The Chameleon} \citep{chameleon_game} and studies a simple but fundamental communication problem: can a model produce a message that is informative to an informed collaborator while remaining ambiguous to an uninformed observer? To measure this, SNEAK evaluates generated messages using two complementary behavioral criteria: \emph{utility}, which captures how well a message communicates to an ally, and \emph{leakage}, which captures how much information it reveals to an observer without access to the secret. Figure~\ref{fig:main} provides an overview of the benchmark setup and evaluation framework. Using this framework, we analyze how modern language models navigate the trade-off between informativeness and identifiability. Our results show that current models often struggle to balance these objectives: messages that are highly useful to collaborators also tend to reveal substantial information more broadly. Under the same evaluation protocol, human participants achieve up to 4$\times$ higher scores than some evaluated LLMs, highlighting a substantial remaining gap in selective communication ability. At the same time, the benchmark exhibits significant headroom, suggesting that controlled information sharing remains a challenging and underdeveloped capability for current systems. The dataset and evaluation scripts will be released upon publication.

Our contributions are as follows:
\begin{itemize}
\item We introduce \textbf{SNEAK}, a benchmark for evaluating selective information sharing in language models, together with a dataset of communication games spanning diverse semantic domains.
\item We propose a behavioral evaluation framework that decomposes communication into \emph{utility} and \emph{leakage}, enabling objective, reproducible evaluation.

\item We show that modern LLMs achieve high utility but struggle with information leakage, exhibiting a utility--leakage trade-off and falling short of human performance.

\end{itemize}

\section{The SNEAK benchmark}

\subsection{Task definition}


SNEAK is an asymmetric communication task designed to evaluate selective information sharing in LLMs. Inspired by the party game \textit{The Chameleon} \citep{chameleon_game}, a model must generate a message about a secret word such that collaborators who know the secret can recognize that the model knows it, while adversaries who do not know the secret cannot reliably infer it. To enable controlled evaluation, SNEAK abstracts this interaction into a one-shot setting with a single generated message and a set of decoys, allowing for consistent and reproducible measurement of utility and leakage.

\paragraph{Game setup}

Each benchmark instance consists of a semantic category $c$, a set of words drawn from that category
\[
W = \{w_1, \dots, w_{|W|}\},
\]
and a secret word $w_s \in W$. A message-generating model $f_\theta$ observes $(c, W, w_s)$ and produces a short natural language message 
\[
m_\theta = f_\theta(c, W, w_s).
\]

To evaluate $m_\theta$, we consider two agents with different information states.

The \textbf{chameleon} (adversary) does not know the secret. Given $(c, W, m_\theta)$, it predicts a distribution over candidate words:
\[
P_{\text{cham}}(w \mid c, W, m_\theta), \quad w \in W.
\]
This distribution represents the adversary's belief about the secret.

The \textbf{ally} (collaborator) knows the secret word $w_s$ and must identify which message corresponds to it. The ally is given $(c, W, w_s)$ and a set of messages
\[
M = \{ m_\theta \} \cup M_{\text{decoy}},
\]
where $M_{\text{decoy}}$ is a set of decoy messages that are generated without access to the secret. The ally predicts a distribution over messages:
\[
P_{\text{ally}}(m \mid c, W, w_s, M), \quad m \in M.
\]
\paragraph{Metrics}
\label{sec:definitions}

Selective communication requires balancing two competing objectives: a message should be recognizable to collaborators who know the secret, but difficult for an uninformed observer to decode. For a generated message $m_\theta$, we define
\[
p_{\text{ally}} = P_{\text{ally}}(m_\theta \mid c, W, w_s, M),
\quad
p_{\text{cham}} = P_{\text{cham}}(w_s \mid c, W, m_\theta).
\]

To account for varying set sizes, we normalize both quantities relative to a random-guess baseline. Let $|M|$ denote the number of candidate messages and $|W|$ the number of candidate words. We define
\[
\tilde{U}(m) = \frac{p_{\text{ally}} - \frac{1}{|M|}}{1 - \frac{1}{|M|}},
\quad
\tilde{L}(m) = \frac{p_{\text{cham}} - \frac{1}{|W|}}{1 - \frac{1}{|W|}}.
\]

These normalized quantities equal $0$ at chance performance and $1$ at perfect performance, but can become negative when performance falls below chance. We clip both quantities at zero, treating below-chance performance as failure and ensuring that all metrics remain interpretable in $[0,1]$:
\[
\text{Utility}(m) = \max(0, \tilde{U}(m)),
\quad
\text{Leakage}(m) = \max(0, \tilde{L}(m)).
\]

We then define a graded score capturing the trade-off between informativeness and secrecy:
\[
\text{SoftScore}(m) = \text{Utility}(m)\cdot\bigl(1 - \text{Leakage}(m)\bigr).
\]

\begin{figure}
    \centering
    \includegraphics[width=0.75\linewidth]{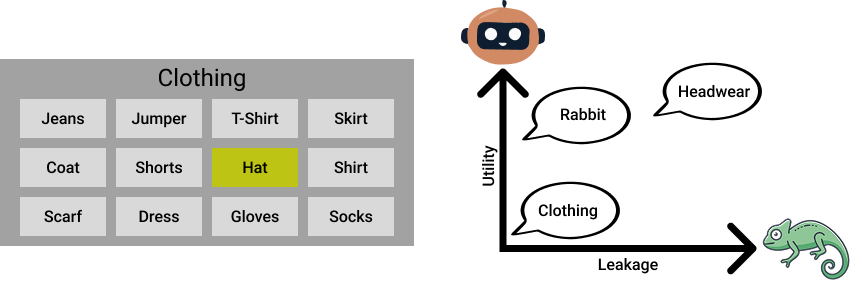}
    \caption{Utility–leakage trade-off for a single instance (secret: hat). The message \emph{rabbit} (as in \emph{a rabbit in a hat}) achieves high utility while maintaining low leakage.}
    \label{fig:example_clues}
\end{figure}

This multiplicative form rewards messages that are both useful and minimally revealing, penalizing failures in either objective. Let $\mathcal{M}_{\leq 5}$ denote the set of all valid natural-language messages of length at most 5 words. For a fixed benchmark instance $(c, W, w_s)$, an optimal message under this objective is
\[
m^\star = \arg\max_{m \in \mathcal{M}_{\leq 5}} \text{SoftScore}(m).
\]
Figure~\ref{fig:example_clues} illustrates this trade-off on a single example, showing how an associative message can remain helpful to an ally without making the secret too easy for the chameleon to identify. All reported values are scaled to percentages by multiplying by $100$, so that all metrics lie in $[0,100]$. In addition, we report a stricter BinaryScore metric. A message is counted as successful if the ally ranks $m_\theta$ highest among the candidate messages and the chameleon does not rank the secret word highest among the candidate words:
\[
\text{BinaryScore}(m_\theta)=
\mathbf{1}\!\left[
m_\theta \in \arg\max_{m \in M} P_{\text{ally}}(m)
\;\land\;
w_s \notin \arg\max_{w \in W} P_{\text{cham}}(w)
\right].
\]

\paragraph{Benchmark objective}

Let $\mathcal{D}$ denote the distribution of benchmark instances $(c, W, w_s)$. For a message-generating model $f_\theta$, we report the average utility, leakage, SoftScore, and BinaryScore across instances:
\[
\mathbb{E}_{(c,W,w_s)\sim\mathcal{D}}
\left[
\text{SoftScore}\bigl(f_\theta(c,W,w_s)\bigr)
\right],
\qquad
\mathbb{E}_{(c,W,w_s)\sim\mathcal{D}}
\left[
\text{BinaryScore}\bigl(f_\theta(c,W,w_s)\bigr)
\right].
\]

In practice, these expectations are approximated by averaging over all benchmark instances.

\subsection{Dataset}


\paragraph{Category sources}

Categories and word instances are derived from the category production norms of \citep{banks2023category}. This dataset contains 117 semantic categories (67 concrete and 50 abstract). In the original study, participants were asked to list as many category members as possible within a fixed time window, producing a large set of human-generated category exemplars. To construct consistent word sets across categories, we select the 12 most frequently produced category members from the category norms dataset. This procedure ensures that word sets contain prototypical and widely recognized category members while avoiding extremely rare or idiosyncratic responses. Treating each candidate word as a possible secret yields 1,394 benchmark instances overall (Table~\ref{tab:dataset_stats}).

\paragraph{Decoy message generation}

To construct the ally evaluation task, we generate 5 decoy messages that are semantically related to the category but are produced without access to the secret word. We query WordNet \citep{miller1995wordnet} to extract lexical attributes and related concepts for each word in the category, aggregate these attributes across words, and retain the most frequent ones as decoy messages. Because these messages are derived solely from the category and word set, they do not encode knowledge of the secret and are therefore treated as incorrect messages in the ally task.

\begin{table}[h]
\centering
\small
\setlength{\tabcolsep}{5pt}
\begin{tabular}{cccc}
\toprule
\textbf{Instances} & \textbf{Categories} & \textbf{Words per Category} & \textbf{Decoys} \\
\midrule
1,394 & 117 & 10--12 & 5 \\
\bottomrule
\end{tabular}
\caption{Summary of the SNEAK dataset. Each instance corresponds to a (category, word set, secret) tuple with associated decoy messages.}
\label{tab:dataset_stats}
\end{table}

\subsection{Evaluation protocol}
\label{sec:evaluation}

We estimate utility and leakage using a behavioral evaluation framework based on simulated agents, framing evaluation as a constrained prediction task rather than subjective LLM-as-judge ratings \citep{li2025generation}. This enables objective and reproducible scoring by reducing evaluation to discrete identification decisions. To estimate leakage, a chameleon agent observes $(c, W, m)$ and predicts a distribution over candidate words, $P_{\text{cham}}(w \mid c, W, m)$; the probability assigned to the secret $w_s$ defines leakage. To estimate utility, an ally agent observes $(c, W, w_s, M)$ and predicts a distribution over messages, $P_{\text{ally}}(m \mid c, W, w_s, M)$; the probability assigned to $m_\theta$ defines utility. Both distributions are obtained via constrained decoding over answer labels (e.g., A, B, C), where probabilities are computed by normalizing first-token logits over the full set of candidate labels. To mitigate label-token biases, we randomize the order of answer choices. Full prompts are provided in Appendix~\ref{sec:prompts}.

\section{Experiments}

We evaluate instruction-tuned language models as message generators on SNEAK. For each benchmark instance, a model receives the category, candidate set, and secret word, and produces a single natural-language message, which is then scored using the utility and leakage metrics from Section~\ref{sec:evaluation}.

\subsection{Reference baselines}

We compare model-generated messages to three simple baselines representing different points on the utility--leakage spectrum: \emph{Random word}, which samples an unrelated WordNet word and is typically uninformative to both agents; \emph{Category-level message}, which uses a synonym or closely related lexicalization of the category name and provides broad semantic information with limited evidence about the secret; and \emph{Secret-level message}, which uses a synonym of the secret word and is highly informative but also highly revealing.

\subsection{Inference-time scaling methods}

To study the effect of additional test-time computation, we apply two standard inference-time scaling methods to each base model: \emph{Chain-of-Thought (CoT)}, which elicits a brief reasoning trace before producing a message \citep{wei2022chain}, and \emph{Self-Enhanced Test-Time Scaling (SETS)}, which generates multiple candidate messages and applies self-verification and self-correction to select a final output \citep{chen2025sets}. Additionally, we introduce a simple two-step transformation method, \emph{Recursive Message Refinement (RMR)}, which first generates a message from the secret and then produces a second message conditioned on the initial message. Intuitively, this procedure encourages more indirect or abstract descriptions, which may help reduce information leakage while preserving signals that allow the ally to recover the intended meaning.

\subsection{Models}

We evaluate a diverse set of instruction-tuned large language models spanning both open- and closed-source families: GPT, Claude, Gemini, DeepSeek, Llama, Gemma, Mixtral, and Qwen \citep{singh2025openai, anthropic2026opus46, team2024gemini, liu2024deepseek, grattafiori2024llama, team2025kimi, jiang2024mixtral, yang2025qwen3}. All models are used as message generators under the same prompting setup. To evaluate generated messages, we use three independent evaluator models (Mixtral-8x7B, Llama-3.3-70B-Instruct, and Qwen-2.5-72B-Instruct), each of which simulates the ally and chameleon agents. Final utility, leakage, and score values are obtained by averaging across evaluators, reducing dependence on any single model and improving robustness. Per-evaluator results are provided in Appendix~\ref{sec:extra_results}.

\begin{table}[t]
\centering

\begin{minipage}[t]{0.48\linewidth}
\centering
\small
\setlength{\tabcolsep}{4pt}

\begin{tabular}{lcccc}
\toprule
Method & Util. & Leak. & S.S. & B.S. \\
\midrule
\rowcolor{lightlightgray}
\multicolumn{5}{l}{\textbf{Reference Baselines}} \\

Random word & 7.85 & 7.89 & 6.71 & 7.34 \\
Category-Synonym & 31.49 & 7.56 & 28.43 & 30.08 \\
Secret-Synonym & 75.73 & 71.02 & 12.31 & 12.15 \\
\midrule
\textbf{Human Participants} & \textbf{91.26} & \textbf{33.81} & \textbf{59.22} & \textbf{54.70}  \\
\midrule
\rowcolor{lightlightgray}
\multicolumn{5}{l}{\textbf{Language Models}} \\

GPT-5.4 & 63.49 & 49.88 & 21.00 & 20.88 \\
Claude Opus 4.6 & 88.41 & 69.20 & 22.23 & 21.43 \\
Gemini 3.1 Pro & 15.32 & 11.85 & 9.54 & 10.04 \\
DeepSeek-V3.2 & 98.52 & 75.42 & 23.81 & 23.17 \\
Llama-4 Scout & 95.18 & 67.11 & 29.48 & 28.79 \\
Qwen3-235B & 90.79 & 67.78 & 25.59 & 24.75 \\
Gemma-3-27B & 92.23 & 59.23 & 34.87 & 33.76 \\
Mixtral-8x22B & 96.53 & 71.00 & 26.63 & 26.02 \\
GPT-OSS-120B & 71.70 & 56.71 & 18.30 & 17.77 \\
\bottomrule
\end{tabular}

\captionof{table}{Main SNEAK benchmark results. We report Utility, Leakage, SoftScore (S.S.), and BinaryScore (B.S.) for reference baselines and language models. Scores are averaged across Mistral, Llama, and Qwen evaluators. Higher utility and lower leakage are desirable; SoftScore captures the trade-off between the two.}
\label{tab:main_results}

\end{minipage}
\hfill
\begin{minipage}[t]{0.48\linewidth}
\centering
\small

\begin{tabular}{lcccc}
\toprule
Method & Util. & Leak. & S.S. & B.S. \\
\midrule

\rowcolor{lightlightgray}
\multicolumn{5}{l}{\textbf{GPT-5.4}} \\
Zero-shot & 63.49 & 49.88 & 21.00 & 20.88 \\
CoT & 84.13 & 67.81 & 21.84 & 21.23 \\
SETS & 87.02 & 68.80 & 23.22 & 22.69 \\
RMR & 84.62 & 68.21 & 22.58 & 22.02 \\

\midrule

\rowcolor{lightlightgray}
\multicolumn{5}{l}{\textbf{Qwen3-235B}} \\
Zero-shot & 90.79 & 67.78 & 25.59 & 24.75 \\
CoT & 92.00 & 67.24 & 26.86 & 25.99 \\
SETS & 91.89 & 69.82 & 24.50 & 23.72 \\
RMR & 77.06 & 47.04 & 33.55 & 32.83 \\

\midrule

\rowcolor{lightlightgray}
\multicolumn{5}{l}{\textbf{Gemma-3-27B}} \\
Zero-shot & 92.23 & 59.23 & 34.87 & 33.76 \\
CoT & 83.43 & 47.94 & 39.21 & 38.38 \\
SETS & 92.03 & 59.88 & 34.32 & 33.48 \\
RMR & 63.28 & 26.42 & 40.80 & 40.91 \\

\bottomrule
\end{tabular}
\captionof{table}{Effect of inference-time scaling methods on SNEAK performance. We compare zero-shot prompting with Chain-of-Thought (CoT), Self-Enhanced Test-Time Scaling (SETS), and Recursive Message Refinement (RMR). Metrics (Utility, Leakage, SoftScore, BinaryScore) are averaged across Mistral, Llama, and Qwen evaluators.}

\label{tab:scaling}

\end{minipage}

\end{table}

\subsection{Main results}

Tables~\ref{tab:main_results} and \ref{tab:scaling} reports average utility, leakage, and composite scores across models and baselines. Overall, modern language models achieve high utility but also incur substantial leakage, indicating a persistent inability to control what information is revealed. Several models (e.g., DeepSeek-V3.2, Mixtral-8x22B, and Llama-4 Scout) approach near-ceiling utility, but do so at the cost of high leakage, placing them in the upper-right region of the utility–leakage plane. In contrast, humans achieve both high utility and substantially lower leakage, resulting in significantly higher SoftScore and BinaryScore. This gap highlights that current models fail to match human-like selective communication, even when they are capable of producing highly informative messages.


\paragraph{Utility--leakage trade-off}

\begin{figure}[t]
    \centering
    \includegraphics[width=0.95\linewidth]{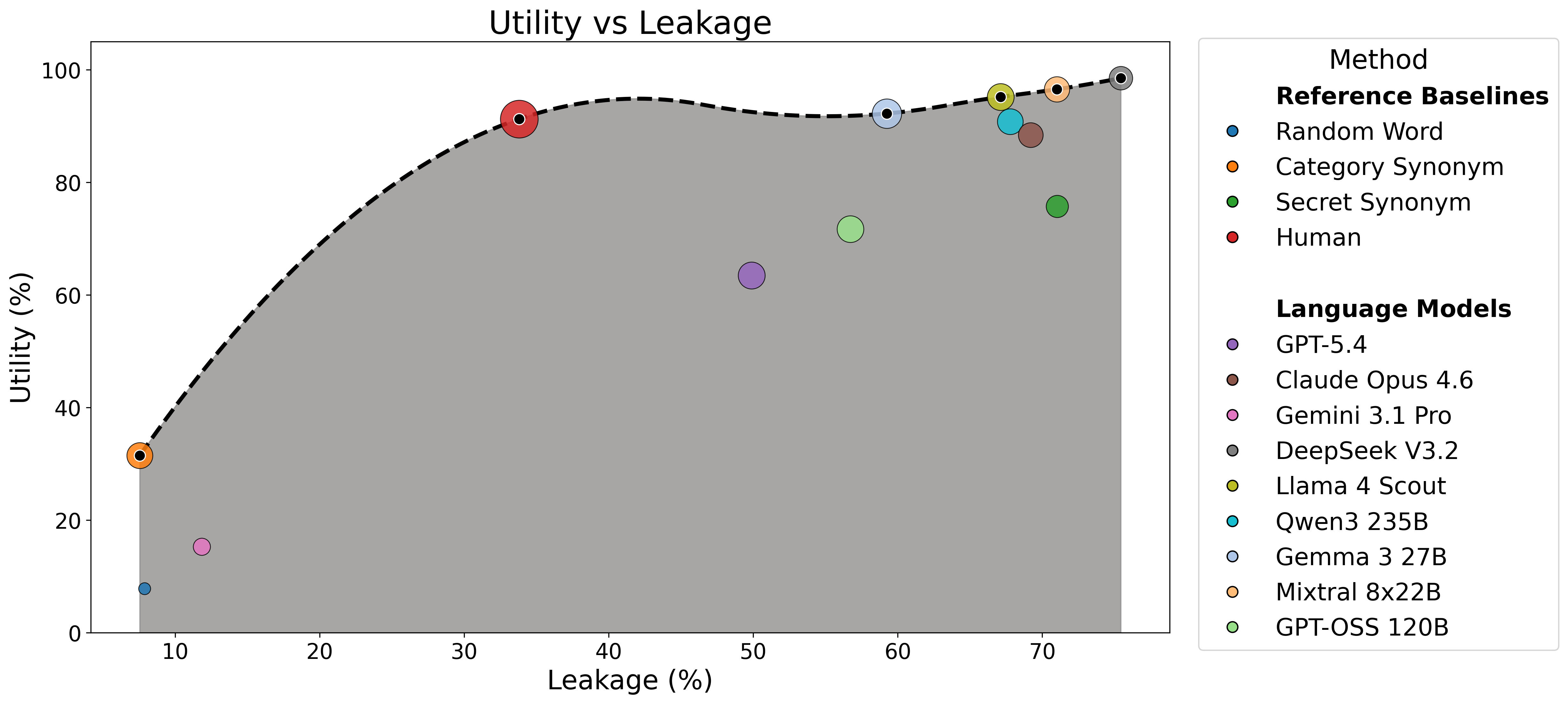}
\caption{Utility--leakage trade-off across message-generating models. Each point corresponds to a baseline, with marker size proportional to SoftScore. The dashed curve shows the empirical Pareto frontier. The upper-left region corresponds to desirable behavior: high utility for the ally and low leakage to the chameleon.}    \label{fig:utility_leakage_tradeoff}
\end{figure}

To better understand model behavior, Figure~\ref{fig:utility_leakage_tradeoff} visualizes the relationship between utility and leakage. Each point corresponds to a baseline, positioned according to its average utility and leakage scores. Across models, we observe a clear trade-off between the two objectives: messages that strongly identify the secret for the ally also tend to increase the probability that the chameleon can infer the secret.  Intentionally revealing messages, such as secret-level lexicalizations, achieve high utility but also high leakage, whereas generic category-level messages reduce leakage at the cost of utility. Modern language models typically occupy intermediate points along this spectrum. The dashed curve shows the empirical Pareto frontier, representing the best observed trade-offs between utility and leakage. Moving this frontier outward—toward the upper-left region—corresponds to achieving higher utility at lower levels of leakage. While some models approach this frontier, none consistently attain strong utility with low leakage, indicating that strategic communication under asymmetric information remains challenging for current systems.

\section{Analysis}

\subsection{Evaluator robustness}

\paragraph{Different evaluator models}

To ensure that benchmark results are not dependent on a single evaluator, we compute utility and leakage using three independent evaluation models: Mixtral-8x7B, Llama-3.3-70B-Instruct, and Qwen-2.5-72B-Instruct. Each evaluator performs the ally and chameleon inference tasks described in Section~\ref{sec:evaluation}, and final benchmark scores are obtained by averaging across evaluators. Across all experiments, score variance is modest: the average standard deviation across evaluators is 5.42 and 4.76 percentage points for utility and leakage respectively, indicating stable absolute scores. Rankings are also highly consistent across evaluators, with Spearman correlations ranging from 0.9042 to 0.9680 for utility, 0.9547 to 0.9929 for leakage (lower is better), and 0.8995 to 0.9547 for the combined SoftScore, indicating strong agreement in model ordering. No single evaluator consistently assigns higher or lower scores across models. These results indicate that the SNEAK evaluation protocol is robust to the choice of evaluator model. We further test for same-model evaluator bias by evaluating outputs from each model using all three evaluators; results remain consistent, with full analysis provided in Appendix~\ref{app:self_bias}.

\paragraph{Human alignment}

Because the SNEAK evaluation relies on simulated agents implemented using language models, we conduct a human study to validate both the task formulation and the alignment of model-based evaluators with human judgments. We sample 117 benchmark instances (one per category) and collect five independent annotations per item for both the ally and chameleon tasks. Humans perform well above chance on both tasks: ally accuracy is 77.3\% at the worker level (chance: 16.7\%) and 88.9\% under plurality voting, while chameleon accuracy is 44.6\% (chance: 8.3\%) and 68.4\% under plurality voting. Inter-annotator agreement is substantial for the ally task (Fleiss’ $\kappa = 0.73$) and moderate for the chameleon task ($\kappa = 0.27$), reflecting greater ambiguity in adversarial inference. We next compare human judgments to LLM-based evaluators. Averaged across three evaluators, model predictions agree with human plurality judgments on 88.9\% of ally items and 66.7\% of chameleon items, increasing to 91.7\% and 74.2\% respectively when restricting to items with a strict human majority. Evaluator probabilities also correlate with human vote shares (ally utility: $r = 0.61$; chameleon leakage: $r = 0.39$), indicating alignment at the distribution level. These results suggest that the model-based evaluation framework provides a reasonable proxy for human judgments while enabling scalable measurement. See Appendix ~\ref{app:human_study} for more details.



\subsection{Semantic structure of category difficulty}
To better understand category-level variation, we analyze difficulty as a function of semantic domain and candidate-space structure. We first split categories into abstract and concrete domains following the source norms. Abstract categories are easier overall, with higher utility (61.8 vs.\ 58.8), lower leakage (47.6 vs.\ 50.0), and a higher mean SoftScore (28.4 vs.\ 25.2). This suggests that abstract domains often permit more indirect associative messages, allowing models to signal knowledge of the secret without making it uniquely identifiable. We also quantify semantic clustering within each category using the average pairwise WordNet similarity among candidate words. Across all categories, higher semantic similarity is associated with lower utility ($r=-0.43$) and lower mean score ($r=-0.28$), indicating that densely clustered candidate sets are harder for selective communication. Intuitively, when candidates are too semantically similar, messages often fail to distinguish the intended secret even for an informed ally. Interestingly, this effect differs by domain: in abstract categories, similarity is more strongly associated with reduced utility and score, whereas in concrete categories it is associated with lower leakage. This suggests that semantic clustering affects strategic communication through different mechanisms in different domains.

\subsection{Sensitivity to candidate and decoy set sizes}

\begin{figure}[h]
    \centering
    \includegraphics[width=0.99\linewidth]{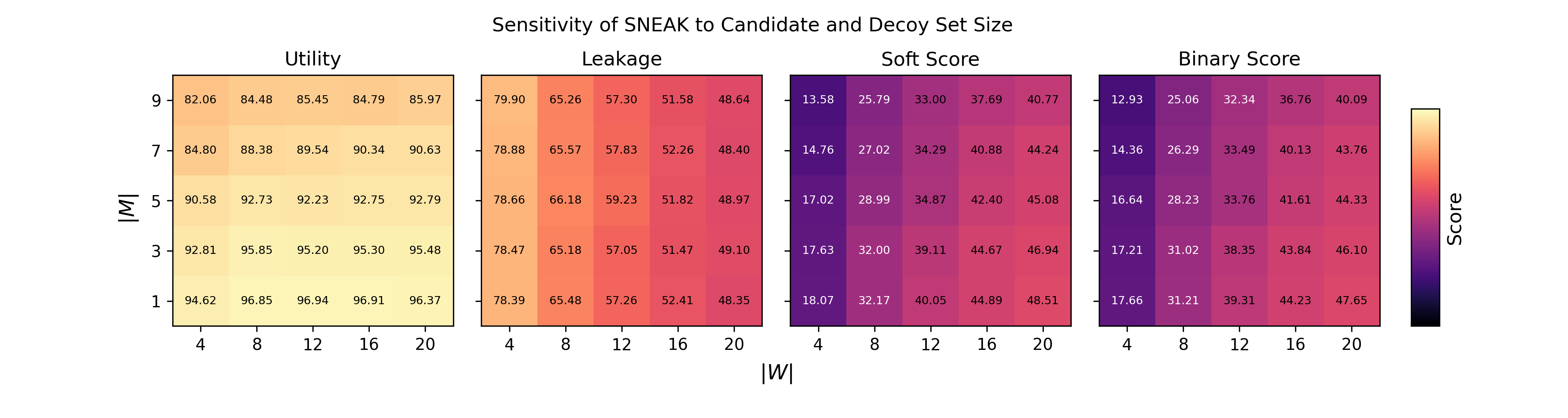}
    \caption{Sensitivity of SNEAK performance for Gemma-3-27B to candidate set size ($|W|$) and number of decoy messages ($|M|$)}
    \label{fig:size_sweep}
\end{figure}

The difficulty of the SNEAK task is governed by two interpretable parameters: the candidate set size $|W|$ and the number of decoy messages $|M|$. Increasing $|W|$ expands the adversary's search space and generally reduces leakage, while increasing $|M|$ makes the ally task harder by requiring identification of the correct message among more alternatives, thereby reducing utility.  To assess the sensitivity of the benchmark to these design choices, we sweep over both parameters and recompute utility, leakage, SoftScore, and BinaryScore using the protocol described in Section~\ref{sec:evaluation}. Figure~\ref{fig:size_sweep} shows the resulting trends for Gemma-3-27B, averaged across evaluator models; results for additional generators are reported in Appendix~\ref{app:sensitivity_set}. The two parameters exert opposing effects on the benchmark: larger candidate sets primarily decrease leakage, whereas more decoy messages primarily decrease utility. As a result, SoftScore and BinaryScore vary in predictable ways across configurations. These results show that SNEAK provides a stable and controllable evaluation setting for selective communication. We further evaluate robustness to decoy construction in Appendix~\ref{sec:decoy_ablation}, finding that results are consistent across alternative decoy generation strategies.

\section{Related work}

\begin{table}[h]
\centering
\small
\resizebox{\linewidth}{!}{
\begin{tabular}{lccccc}
\toprule
Benchmark & 
\makecell{Multi-Agent?} & 
\makecell{Asymmetric\\Info?} & 
\makecell{Strategic\\Signaling?} & 
\makecell{Utterance\\Eval?} & 
\makecell{Leakage\\Measured?} \\
\midrule

MMLU-Pro \citep{wang2024mmlu} 
& \textcolor{red}{\ding{55}} 
& \textcolor{red}{\ding{55}} 
& \textcolor{red}{\ding{55}} 
& \textcolor{green!60!black}{\ding{51}} 
& \textcolor{red}{\ding{55}} \\


MT-Bench \citep{zheng2023judging} 
& \textcolor{red}{\ding{55}} 
& \textcolor{red}{\ding{55}} 
& \textcolor{red}{\ding{55}} 
& \textcolor{green!60!black}{\ding{51}} 
& \textcolor{red}{\ding{55}} \\

SocialMaze \citep{xu2025socialmaze} 
& \textcolor{green!60!black}{\ding{51}} 
& \textcolor{green!60!black}{\ding{51}} 
& \textcolor{red}{\ding{55}} 
& \textcolor{red}{\ding{55}} 
& \textcolor{red}{\ding{55}} \\

Mini-Mafia \citep{costa2025deceive} 
& \textcolor{green!60!black}{\ding{51}} 
& \textcolor{green!60!black}{\ding{51}} 
& \textcolor{green!60!black}{\ding{51}} 
& \textcolor{red}{\ding{55}} 
& \textcolor{red}{\ding{55}} \\

WhoIsSpy \citep{chen2025llmspark} 
& \textcolor{green!60!black}{\ding{51}} 
& \textcolor{green!60!black}{\ding{51}} 
& \textcolor{green!60!black}{\ding{51}} 
& \textcolor{red}{\ding{55}} 
& \textcolor{red}{\ding{55}} \\

\midrule
\textbf{SNEAK (ours)} 
& \textcolor{green!60!black}{\ding{51}} 
& \textcolor{green!60!black}{\ding{51}} 
& \textcolor{green!60!black}{\ding{51}} 
& \textcolor{green!60!black}{\ding{51}} 
& \textcolor{green!60!black}{\ding{51}} \\
\bottomrule
\end{tabular}
}
\caption{Comparison of SNEAK with representative LLM benchmarks across key properties.}
\label{tab:benchmark_comparison}
\end{table}

A large body of work evaluates language models on factual knowledge, reasoning, and dialogue capabilities. Benchmarks such as MMLU \citep{hendrycks2020measuring} and GSM8K \citep{cobbe2021training} measure question answering and mathematical reasoning, while MT-Bench \citep{zheng2023judging} evaluates conversational helpfulness using model-based judges. More broadly, recent benchmark suites assess capabilities such as coding, planning, and instruction following \citep{mohammadi2025evaluation}. These benchmarks provide \emph{utterance-level evaluation} of model outputs, but they generally assume a single agent with full access to the task information. In contrast, recent multi-agent and social-deduction benchmarks (e.g., SocialMaze, Mini-Mafia, WhoIsSpy) introduce asymmetric information and strategic interaction, but typically evaluate behavior through multi-turn game outcomes rather than directly scoring individual utterances. As summarized in Table~\ref{tab:benchmark_comparison}, SNEAK uniquely combines asymmetric information, strategic signaling, and multi-agent interaction with \emph{utterance-level evaluation} in a single-turn setting, and explicitly measures information leakage.

Our benchmark is also closely related to research on pragmatics and reasoning about the beliefs of other agents. Classical accounts of communication emphasize that speakers choose utterances relative to shared context and listener beliefs, often balancing informativeness with ambiguity \citep{grice1975logic}. Computational frameworks such as Rational Speech Acts formalize this process as recursive reasoning between speakers and listeners \citep{frank2012predicting, goodman2016pragmatic}. Related ideas appear in signaling games, emergent communication, and theory-of-mind reasoning in language models \citep{lazaridou2016multi, havrylov2017emergence, kosinski2023theory, yao2023react, sap2019social}. While these lines of work study belief inference, pragmatic reasoning, and strategic interaction, they do not directly evaluate whether models can produce utterances that selectively reveal information to different listeners. SNEAK operationalizes this as a controlled one-shot setting where models inform collaborators while limiting inference by uninformed observers.

\section{Limitations}

SNEAK evaluates selective information sharing in a simplified communication setting. The benchmark focuses on single-turn message generation and does not capture the richer dynamics of multi-turn interaction, discussion, or deception present in real communication. Evaluation relies on simulated ally and adversary agents implemented using LLMs. While this enables scalable evaluation, model-based judges may not perfectly reflect how humans interpret messages. Finally, the dataset is derived from semantic category norms and therefore emphasizes structured conceptual domains. Real-world communication often involves richer context and background knowledge that are not represented in this setting. We view SNEAK as an initial step toward evaluating selective communication in LLMs and hope future work will extend this framework to more realistic multi-agent settings.
\section{Conclusion}

We introduced \textbf{SNEAK}, a benchmark for evaluating selective information sharing in LLMs. The benchmark measures whether models can generate messages that are informative to collaborators who share context while remaining ambiguous to observers who do not. We propose a behavioral framework that quantifies both the utility of a message for allies and the information leakage to adversaries. Our experiments show that modern LLMs often struggle to balance these competing objectives: messages that are highly informative for collaborators frequently reveal substantial information to adversaries. This suggests that strategic communication under asymmetric information remains a challenging capability for current systems. By providing a controlled setting for measuring how language models manage information flow between agents, SNEAK highlights selective communication as an important and underexplored dimension of language model behavior.


\section*{Ethics Statement}

This work studies selective information sharing in language models, focusing on how agents can communicate useful information to intended recipients while limiting unintended inference. While this capability has beneficial applications (e.g., tutoring, privacy-preserving communication, and controlled information disclosure), it may also raise concerns related to strategic obfuscation or misuse in deceptive settings. Our goal is to provide a controlled evaluation framework for measuring these behaviors, rather than to promote or optimize for deceptive communication. The SNEAK benchmark isolates a simplified, single-turn setting and does not model real-world deployment contexts or incentives. We emphasize that the ability to control information leakage should be studied alongside considerations of transparency, safety, and appropriate use. We release our dataset and evaluation framework to support further research on responsible communication in AI systems, including improving models’ ability to balance informativeness with privacy and safety considerations.

\section*{Acknowledgments}
We thank Amir Avsian for introducing us to the game \textit{The Chameleon}, which inspired the formulation of this task.

\bibliography{colm2026_conference}

@incollection{grice1975logic,
  title={Logic and conversation},
  author={Grice, Herbert P},
  booktitle={Speech acts},
  pages={41--58},
  year={1975},
  publisher={Brill}
}

@article{frank2012predicting,
  title={Predicting pragmatic reasoning in language games},
  author={Frank, Michael C. and Goodman, Noah D.},
  journal={Science},
  volume={336},
  number={6084},
  pages={998},
  year={2012}
}

@article{goodman2016pragmatic,
  title={Pragmatic language interpretation as probabilistic inference},
  author={Goodman, Noah D. and Frank, Michael C.},
  journal={Trends in Cognitive Sciences},
  volume={20},
  number={11},
  pages={818--829},
  year={2016}
}

@inproceedings{andreas2022language,
  title={Language models as agent models},
  author={Andreas, Jacob},
  booktitle={Findings of the Association for Computational Linguistics: EMNLP 2022},
  pages={5769--5779},
  year={2022}
}

@inproceedings{lazaridou2016multi,
  title={Multi-agent cooperation and the emergence of (natural) language},
  author={Lazaridou, Angeliki and Peysakhovich, Alexander and Baroni, Marco},
  booktitle={ICLR},
  year={2017}
}

@inproceedings{havrylov2017emergence,
  title={Emergence of language with multi-agent games},
  author={Havrylov, Serhii and Titov, Ivan},
  booktitle={NeurIPS},
  year={2017}
}

@article{kosinski2023theory,
  title={Theory of mind may have spontaneously emerged in large language models},
  author={Kosinski, Michal},
  journal={arXiv preprint arXiv:2302.02083},
  year={2023}
}

@inproceedings{sap2019social,
  title={Social IQa: Commonsense reasoning about social interactions},
  author={Sap, Maarten and Rashkin, Hannah and Chen, Derek and Le Bras, Ronan and Choi, Yejin},
  booktitle={Proceedings of the 2019 conference on empirical methods in natural language processing and the 9th international joint conference on natural language processing (EMNLP-IJCNLP)},
  pages={4463--4473},
  year={2019}
}

@inproceedings{yao2023react,
  title={ReAct: Synergizing reasoning and acting in language models},
  author={Yao, Shunyu and others},
  booktitle={ICLR},
  year={2023}
}

@article{wang2024survey,
  title={A survey on large language model based autonomous agents},
  author={Wang, Lei and Ma, Chen and Feng, Xueyang and Zhang, Zeyu and Yang, Hao and Zhang, Jingsen and Chen, Zhiyuan and Tang, Jiakai and Chen, Xu and Lin, Yankai and others},
  journal={Frontiers of Computer Science},
  volume={18},
  number={6},
  pages={186345},
  year={2024},
  publisher={Springer}
}

@inproceedings{wu2024autogen,
  title={Autogen: Enabling next-gen LLM applications via multi-agent conversations},
  author={Wu, Qingyun and Bansal, Gagan and Zhang, Jieyu and Wu, Yiran and Li, Beibin and Zhu, Erkang and Jiang, Li and Zhang, Xiaoyun and Zhang, Shaokun and Liu, Jiale and others},
  booktitle={First conference on language modeling},
  year={2024}
}

@inproceedings{park2023generative,
  title={Generative agents: Interactive simulacra of human behavior},
  author={Park, Joon Sung and O'Brien, Joseph and Cai, Carrie Jun and Morris, Meredith Ringel and Liang, Percy and Bernstein, Michael S},
  booktitle={Proceedings of the 36th annual acm symposium on user interface software and technology},
  pages={1--22},
  year={2023}
}

@inproceedings{zhan2024let,
  title={Let’s negotiate! a survey of negotiation dialogue systems},
  author={Zhan, Haolan and Wang, Yufei and Li, Zhuang and Feng, Tao and Hua, Yuncheng and Sharma, Suraj and Qu, Lizhen and Azad, Zhaleh Semnani and Zukerman, Ingrid and Haf, Reza},
  booktitle={Findings of the Association for Computational Linguistics: EACL 2024},
  pages={2019--2031},
  year={2024}
}

@article{meta2022human,
  title={Human-level play in the game of diplomacy by combining language models with strategic reasoning},
  author={{FAIR} and Bakhtin, Anton and Brown, Noam and Dinan, Emily and Farina, Gabriele and Flaherty, Colin and Fried, Daniel and Goff, Andrew and Gray, Jonathan and Hu, Hengyuan and others},
  journal={Science},
  volume={378},
  number={6624},
  pages={1067--1074},
  year={2022},
  publisher={American Association for the Advancement of Science}
}

@inproceedings{mohammadi2025evaluation,
  title={Evaluation and benchmarking of llm agents: A survey},
  author={Mohammadi, Mahmoud and Li, Yipeng and Lo, Jane and Yip, Wendy},
  booktitle={Proceedings of the 31st ACM SIGKDD Conference on Knowledge Discovery and Data Mining V. 2},
  pages={6129--6139},
  year={2025}
}

@article{miller1995wordnet,
  title={WordNet: a lexical database for English},
  author={Miller, George A},
  journal={Communications of the ACM},
  volume={38},
  number={11},
  pages={39--41},
  year={1995},
  publisher={ACM New York, NY, USA}
}

@article{banks2023category,
  title={Category production norms for 117 concrete and abstract categories},
  author={Banks, Briony and Connell, Louise},
  journal={Behavior Research Methods},
  volume={55},
  number={3},
  pages={1292--1313},
  year={2023},
  publisher={Springer}
}

@article{grattafiori2024llama,
  title={The llama 3 herd of models},
  author={Grattafiori, Aaron and Dubey, Abhimanyu and Jauhri, Abhinav and Pandey, Abhinav and Kadian, Abhishek and Al-Dahle, Ahmad and Letman, Aiesha and Mathur, Akhil and Schelten, Alan and Vaughan, Alex and others},
  journal={arXiv preprint arXiv:2407.21783},
  year={2024}
}

@article{team2025kimi,
  title={Kimi-vl technical report},
  author={Team, Kimi and Du, Angang and Yin, Bohong and Xing, Bowei and Qu, Bowen and Wang, Bowen and Chen, Cheng and Zhang, Chenlin and Du, Chenzhuang and Wei, Chu and others},
  journal={arXiv preprint arXiv:2504.07491},
  year={2025}
}

@article{jiang2024mixtral,
  title={Mixtral of experts},
  author={Jiang, Albert Q and Sablayrolles, Alexandre and Roux, Antoine and Mensch, Arthur and Savary, Blanche and Bamford, Chris and Chaplot, Devendra Singh and Casas, Diego de las and Hanna, Emma Bou and Bressand, Florian and others},
  journal={arXiv preprint arXiv:2401.04088},
  year={2024}
}

@article{yang2025qwen3,
  title={Qwen3 technical report},
  author={Yang, An and Li, Anfeng and Yang, Baosong and Zhang, Beichen and Hui, Binyuan and Zheng, Bo and Yu, Bowen and Gao, Chang and Huang, Chengen and Lv, Chenxu and others},
  journal={arXiv preprint arXiv:2505.09388},
  year={2025}
}

@article{liu2024deepseek,
  title={Deepseek-v3 technical report},
  author={Liu, Aixin and Feng, Bei and Xue, Bing and Wang, Bingxuan and Wu, Bochao and Lu, Chengda and Zhao, Chenggang and Deng, Chengqi and Zhang, Chenyu and Ruan, Chong and others},
  journal={arXiv preprint arXiv:2412.19437},
  year={2024}
}

@article{costa2025deceive,
  title={Deceive, Detect, and Disclose: Large Language Models Play Mini-Mafia},
  author={Costa, Davi Bastos and Vicente, Renato},
  journal={arXiv preprint arXiv:2509.23023},
  year={2025}
}

@article{chen2025llmspark,
  title={LLMsPark: A Benchmark for Evaluating Large Language Models in Strategic Gaming Contexts},
  author={Chen, Junhao and Sun, Jingbo and Li, Xiang and Xin, Haidong and Xue, Yuhao and Xu, Yibin and Zhao, Hao},
  journal={arXiv preprint arXiv:2509.16610},
  year={2025}
}

@article{singh2025openai,
  title={Openai gpt-5 system card},
  author={Singh, Aaditya and Fry, Adam and Perelman, Adam and Tart, Adam and Ganesh, Adi and El-Kishky, Ahmed and McLaughlin, Aidan and Low, Aiden and Ostrow, AJ and Ananthram, Akhila and others},
  journal={arXiv preprint arXiv:2601.03267},
  year={2025}
}

@misc{anthropic2026opus46,
  title        = {Claude Opus 4.6 System Card},
  author       = {{Anthropic}},
  year         = {2026},
  howpublished = {\url{https://www-cdn.anthropic.com/14e4fb01875d2a69f646fa5e574dea2b1c0ff7b5.pdf}},
  note         = {Anthropic technical report}
}

@article{team2024gemini,
  title={Gemini 1.5: Unlocking multimodal understanding across millions of tokens of context},
  author={Team, Gemini and Georgiev, Petko and Lei, Ving Ian and Burnell, Ryan and Bai, Libin and Gulati, Anmol and Tanzer, Garrett and Vincent, Damien and Pan, Zhufeng and Wang, Shibo and others},
  journal={arXiv preprint arXiv:2403.05530},
  year={2024}
}

@article{hendrycks2020measuring,
  title={Measuring massive multitask language understanding},
  author={Hendrycks, Dan and Burns, Collin and Basart, Steven and Zou, Andy and Mazeika, Mantas and Song, Dawn and Steinhardt, Jacob},
  journal={arXiv preprint arXiv:2009.03300},
  year={2020}
}

@article{cobbe2021training,
  title={Training verifiers to solve math word problems},
  author={Cobbe, Karl and Kosaraju, Vineet and Bavarian, Mohammad and Chen, Mark and Jun, Heewoo and Kaiser, Lukasz and Plappert, Matthias and Tworek, Jerry and Hilton, Jacob and Nakano, Reiichiro and others},
  journal={arXiv preprint arXiv:2110.14168},
  year={2021}
}

@article{zheng2023judging,
  title={Judging llm-as-a-judge with mt-bench and chatbot arena},
  author={Zheng, Lianmin and Chiang, Wei-Lin and Sheng, Ying and Zhuang, Siyuan and Wu, Zhanghao and Zhuang, Yonghao and Lin, Zi and Li, Zhuohan and Li, Dacheng and Xing, Eric and others},
  journal={Advances in neural information processing systems},
  volume={36},
  pages={46595--46623},
  year={2023}
}

@article{xu2025socialmaze,
  title={Socialmaze: A benchmark for evaluating social reasoning in large language models},
  author={Xu, Zixiang and Wang, Yanbo and Huang, Yue and Ye, Jiayi and Zhuang, Haomin and Song, Zirui and Gao, Lang and Wang, Chenxi and Chen, Zhaorun and Zhou, Yujun and others},
  journal={arXiv preprint arXiv:2505.23713},
  year={2025}
}

@article{wei2022chain,
  title={Chain-of-thought prompting elicits reasoning in large language models},
  author={Wei, Jason and Wang, Xuezhi and Schuurmans, Dale and Bosma, Maarten and Xia, Fei and Chi, Ed and Le, Quoc V and Zhou, Denny and others},
  journal={Advances in neural information processing systems},
  volume={35},
  pages={24824--24837},
  year={2022}
}

@article{chen2025sets,
  title={Sets: Leveraging self-verification and self-correction for improved test-time scaling},
  author={Chen, Jiefeng and Ren, Jie and Chen, Xinyun and Yang, Chengrun and Sun, Ruoxi and Yoon, Jinsung and Ar{\i}k, Sercan {\"O}},
  journal={arXiv preprint arXiv:2501.19306},
  year={2025}
}

@inproceedings{kazemi2025big,
  title={Big-bench extra hard},
  author={Kazemi, Mehran and Fatemi, Bahare and Bansal, Hritik and Palowitch, John and Anastasiou, Chrysovalantis and Mehta, Sanket Vaibhav and Jain, Lalit K and Aglietti, Virginia and Jindal, Disha and Chen, Yuanzhu Peter and others},
  booktitle={Proceedings of the 63rd Annual Meeting of the Association for Computational Linguistics (Volume 1: Long Papers)},
  pages={26473--26501},
  year={2025}
}

@article{zhou2023instruction,
  title={Instruction-following evaluation for large language models},
  author={Zhou, Jeffrey and Lu, Tianjian and Mishra, Swaroop and Brahma, Siddhartha and Basu, Sujoy and Luan, Yi and Zhou, Denny and Hou, Le},
  journal={arXiv preprint arXiv:2311.07911},
  year={2023}
}

@article{phan2025humanity,
  title={Humanity's last exam},
  author={Phan, Long and Gatti, Alice and Han, Ziwen and Li, Nathaniel and Hu, Josephina and Zhang, Hugh and Zhang, Chen Bo Calvin and Shaaban, Mohamed and Ling, John and Shi, Sean and others},
  journal={arXiv preprint arXiv:2501.14249},
  year={2025}
}

@article{du2025supergpqa,
  title={Supergpqa: Scaling llm evaluation across 285 graduate disciplines},
  author={Du, Xinrun and Yao, Yifan and Ma, Kaijing and Wang, Bingli and Zheng, Tianyu and Zhu, King and Liu, Minghao and Liang, Yiming and Jin, Xiaolong and Wei, Zhenlin and others},
  journal={arXiv preprint arXiv:2502.14739},
  year={2025}
}

@article{shafto2014rational,
  title={A rational account of pedagogical reasoning: Teaching by, and learning from, examples},
  author={Shafto, Patrick and Goodman, Noah D and Griffiths, Thomas L},
  journal={Cognitive psychology},
  volume={71},
  pages={55--89},
  year={2014},
  publisher={Elsevier}
}

@article{crawford1982strategic,
  title={Strategic information transmission},
  author={Crawford, Vincent P and Sobel, Joel},
  journal={Econometrica: Journal of the Econometric Society},
  pages={1431--1451},
  year={1982},
  publisher={JSTOR}
}

@inproceedings{li2025generation,
  title={From generation to judgment: Opportunities and challenges of llm-as-a-judge},
  author={Li, Dawei and Jiang, Bohan and Huang, Liangjie and Beigi, Alimohammad and Zhao, Chengshuai and Tan, Zhen and Bhattacharjee, Amrita and Jiang, Yuxuan and Chen, Canyu and Wu, Tianhao and others},
  booktitle={Proceedings of the 2025 Conference on Empirical Methods in Natural Language Processing},
  pages={2757--2791},
  year={2025}
}

@article{wang2024mmlu,
  title={Mmlu-pro: A more robust and challenging multi-task language understanding benchmark},
  author={Wang, Yubo and Ma, Xueguang and Zhang, Ge and Ni, Yuansheng and Chandra, Abhranil and Guo, Shiguang and Ren, Weiming and Arulraj, Aaran and He, Xuan and Jiang, Ziyan and others},
  journal={Advances in Neural Information Processing Systems},
  volume={37},
  pages={95266--95290},
  year={2024}
}

@article{belland2013framework,
  title={A framework for designing scaffolds that improve motivation and cognition},
  author={Belland, Brian R and Kim, ChanMin and Hannafin, Michael J},
  journal={Educational psychologist},
  volume={48},
  number={4},
  pages={243--270},
  year={2013},
  publisher={Taylor \& Francis}
}

@article{ma2014intelligent,
  title={Intelligent tutoring systems and learning outcomes: A meta-analysis.},
  author={Ma, Wenting and Adesope, Olusola O and Nesbit, John C and Liu, Qing},
  journal={Journal of educational psychology},
  volume={106},
  number={4},
  pages={901},
  year={2014},
  publisher={American Psychological Association}
}

@article{mcrae2005semantic,
  title={Semantic feature production norms for a large set of living and nonliving things},
  author={McRae, Ken and Cree, George S and Seidenberg, Mark S and McNorgan, Chris},
  journal={Behavior research methods},
  volume={37},
  number={4},
  pages={547--559},
  year={2005},
  publisher={Springer}
}

@article{wu2009perceptual,
  title={Perceptual simulation in conceptual combination: Evidence from property generation},
  author={Wu, Ling-ling and Barsalou, Lawrence W},
  journal={Acta psychologica},
  volume={132},
  number={2},
  pages={173--189},
  year={2009},
  publisher={Elsevier}
}

@misc{chameleon_game,
  title = {The Chameleon},
  author = {{Big Potato Games}},
  year = {2017},
  howpublished = {\url{https://bigpotato.com/products/the-chameleon}},
  note = {Board game}
}
\bibliographystyle{colm2026_conference}

\clearpage
\appendix

\section{Prompts}
\label{sec:prompts}

This section presents the prompt templates used in the SNEAK benchmark for message generation and evaluation. We include the prompt used by the message-generating model, as well as the prompts used by the evaluator models that simulate the ally and chameleon agents. Variables such as the category, candidate set, secret word, and message set are inserted programmatically at runtime.

\paragraph{Message Generation}\mbox{}\\[.1em]
\par\vspace{-0.5\baselineskip}

\begin{tcolorbox}
\textbf{System:} You are an expert player of the party game ``Chameleon''.

You will be given:
\begin{itemize}
    \item A category
    \item A list of candidate words
    \item The secret word (one of the candidates)
\end{itemize}

Your task is to produce a single message that signals to other honest players that you know the secret word, while not revealing the word directly.

Guidelines:
\begin{itemize}
    \item The message should strongly hint at the secret word.
    \item The message should help distinguish the secret from the other candidates.
    \item The message must not contain the secret word or obvious variations of it.
    \item Do not spell the word, give its initials, or describe its letters.
    \item Do not translate the word or use a direct synonym.
    \item Avoid messages that apply equally well to many of the candidates.
    \item Keep the message short and natural (1-5 words).
\end{itemize}

Think carefully about what property uniquely characterizes the secret compared to the other candidates.

Output format:

\texttt{<message>YOUR message HERE</message>}
\end{tcolorbox}

\vspace{12pt}
\begin{tcolorbox}
\textbf{User:}

\texttt{Category: \{CATEGORY\}}

\texttt{Candidate words: \{CANDIDATES\}}

\texttt{Secret word: \{SECRET\}}

\texttt{Write one message that hints at the secret word without revealing it.}

\texttt{Return the result in the format:}

\texttt{<message>...</message>}
\end{tcolorbox}

\paragraph{Chameleon Evaluation}\mbox{}\\[.1em]
\par\vspace{-0.5\baselineskip}

\begin{tcolorbox}
\textbf{System:} You are the judge for the game ``Chameleon''.

You will be given:
\begin{itemize}
    \item A category
    \item A list of candidate words, each tagged with a letter label
    \item A single message from a player
\end{itemize}

Task:

Infer which labeled candidate is the secret word the message refers to.

Rules:
\begin{itemize}
    \item Output exactly one uppercase letter label.
    \item Do not output any explanation or extra text.
\end{itemize}
\end{tcolorbox}

\vspace{12pt}
\begin{tcolorbox}
\textbf{User:}

\texttt{Category: \{CATEGORY\}}

\texttt{Candidate words:}
\texttt{A) \{WORD\_1\}}
\texttt{B) \{WORD\_2\}}
\texttt{C) \{WORD\_3\}}
\texttt{...}

\texttt{Message:}
\texttt{\{MESSAGE\}}

\texttt{Choose exactly one label from: A, B, C, ...}

\texttt{Answer with only that single letter.}
\end{tcolorbox}

\paragraph{Ally Evaluation}\mbox{}\\[.1em]
\par\vspace{-0.5\baselineskip}

\begin{tcolorbox}
\textbf{System:} You are the Ally judge in the game ``Chameleon''.

You will be given:
\begin{itemize}
    \item A category
    \item A list of candidate words
    \item The true secret word
    \item A set of messages, each tagged with a letter label
\end{itemize}

Task:

Decide which labeled message most likely came from a player who knew the secret word.

Rules:
\begin{itemize}
    \item Output exactly one uppercase letter label.
    \item Do not output any explanation or extra text.
\end{itemize}
\end{tcolorbox}

\vspace{12pt}
\begin{tcolorbox}
\textbf{User:}

\texttt{Category: \{CATEGORY\}}

\texttt{Candidate words:}
\texttt{\{CANDIDATES\}}

\texttt{Secret word:}
\texttt{\{SECRET\}}

\texttt{Messages:}
\texttt{A) message\_0: \{message\_0\}}
\texttt{B) message\_1: \{message\_1\}}
\texttt{C) message\_2: \{message\_2\}}
\texttt{...}

\texttt{Choose exactly one label from: A, B, C, ...}

\texttt{Answer with only that single letter.}
\end{tcolorbox}

\vspace{8pt}
For message generation, we extract the content of the final \texttt{<message>...</message>} span from the model output. If no such span is present, the generation is marked as a parsing failure. For ally and chameleon evaluation, models are prompted to produce a single letter corresponding to one of the presented options. At evaluation time, we use constrained label scoring to obtain a probability distribution over the available letter labels, which is then mapped back to the corresponding candidate words or message identifiers. These probabilities are used to compute utility and leakage. Outputs that cannot be parsed or do not assign valid probabilities to the full label set are marked as evaluation parsing failures.

\section{Human Study}
\label{app:human_study}

We conduct human studies to evaluate both (i) human performance in message generation and (ii) human judgments in message evaluation, as well as agreement between human annotators and LLM-based evaluators. To ensure consistent coverage across semantic domains, both studies use the same set of 117 benchmark instances (one per category), but employ separate annotation setups.

\paragraph{Annotators.}
All annotations were collected on Amazon Mechanical Turk (AMT) using workers with a minimum approval rate of 95\%. Annotators were restricted to English-speaking countries to ensure reliable comprehension. Compensation was set to meet or exceed the U.S.\ federal minimum wage based on conservative task-time estimates. No personally identifiable information was collected.

\subsection{Human Message Generation}

We evaluate human performance on the message generation task using the shared set of 117 benchmark instances. For each instance, annotators are given the category, candidate word set, and secret word, and are asked to produce a single-word message that indicates knowledge of the secret without revealing it too directly. Each instance is annotated by three independent annotators.

To assess performance, we evaluate human-generated messages using the automated evaluation protocol described in Section~\ref{sec:evaluation}, measuring utility, leakage, SoftScore, and BinaryScore. For each instance, we collect three independent responses and report the highest-scoring response, approximating an upper bound on human performance under annotator variability. Results are then averaged across instances.

\begin{table}[h]
\centering
\begin{tabular}{lcccc}
\toprule
\textbf{Source} & \textbf{Utility} & \textbf{Leakage} & \textbf{SoftScore} & \textbf{BinaryScore} \\
\midrule
Human Participants & 91.26 & 33.81 & 59.22 & 54.70 \\
\bottomrule
\end{tabular}
\caption{Performance of human-generated messages under the SNEAK evaluation protocol. Each instance includes three human responses; we report the best-performing response per instance.}
\label{tab:human_generation}
\end{table}

\subsection{Human Message Evaluation}

We evaluate human judgments on the message evaluation tasks using the same set of 117 benchmark instances, but with a separate annotation setup. Each item is annotated independently by five annotators.

\textbf{Ally evaluation:} given the category, candidate set, secret word, and a set of messages (one gold and several decoys), select the message corresponding to the secret.

\textbf{Chameleon evaluation:} given the category, candidate set, and a message, select the most likely secret word.

We report both worker-level accuracy and item-level accuracy obtained via plurality voting across annotators.

\begin{table}[h]
\centering
\begin{tabular}{lcc}
\toprule
\textbf{Task} & \textbf{Worker Acc.} & \textbf{Plurality Acc.} \\
\midrule
Ally & 77.3\% & 88.9\% \\
Chameleon & 44.6\% & 68.4\% \\
\bottomrule
\end{tabular}
\caption{Human performance on SNEAK message evaluation tasks. Chance levels are 16.7\% for ally and 8.3\% for chameleon.}
\label{tab:human_evaluation}
\end{table}

Inter-annotator agreement is substantial for the ally task (Fleiss’ $\kappa = 0.73$) and moderate for the chameleon task ($\kappa = 0.27$), reflecting greater ambiguity in adversarial inference.

\subsection{Human--LLM Agreement}

We compare human judgments in message evaluation to LLM-based evaluators at the item level. Human decisions are aggregated using plurality voting across five annotators, while LLM predictions are obtained by averaging across three evaluator models (Mixtral-8x7B-Instruct-v0.1, Llama-3.3-70B-Instruct, and Qwen-2.5-72B-Instruct).

\begin{table}[h]
\centering
\begin{tabular}{lcc}
\toprule
\textbf{Task} & \textbf{Top-1 Agreement} & \textbf{Strict-Majority Agreement} \\
\midrule
Ally & 88.9\% & 91.7\% \\
Chameleon & 66.7\% & 74.2\% \\
\bottomrule
\end{tabular}
\caption{Agreement between LLM evaluators and human plurality judgments. Strict-majority considers only items with a clear human majority.}
\label{tab:human_llm_agreement}
\end{table}

To assess alignment at the distribution level, we compare human vote shares with evaluator probabilities. The correlation between human and model estimates is $r = 0.61$ for ally utility and $r = 0.39$ for chameleon leakage, indicating moderate alignment between human judgments and model-based evaluation.

These results support the use of LLM-based evaluators as scalable proxies for human judgments in the SNEAK benchmark.

\subsection{Mturk Interfaces}

\begin{figure}[h]
\centering
\includegraphics[width=0.9\columnwidth]{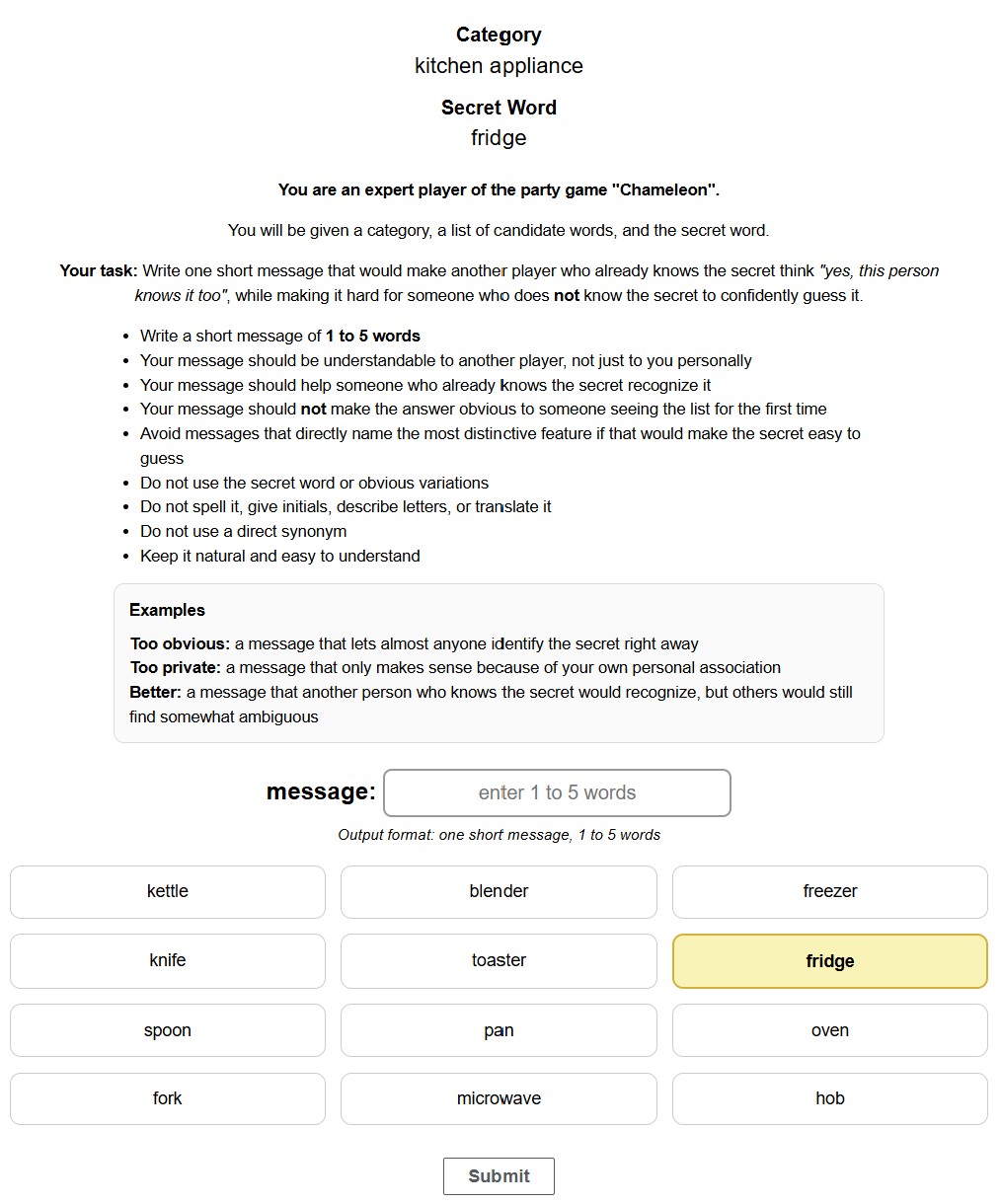}
\caption{
Human annotation interface for message generation. Annotators are given the category, candidate set, and secret word, and are asked to produce a 1-5 word message that signals knowledge of the secret without revealing it directly.
}
\label{fig:human_generation_ui}
\end{figure}

\begin{figure}
    \centering
    \includegraphics[width=0.95\linewidth]{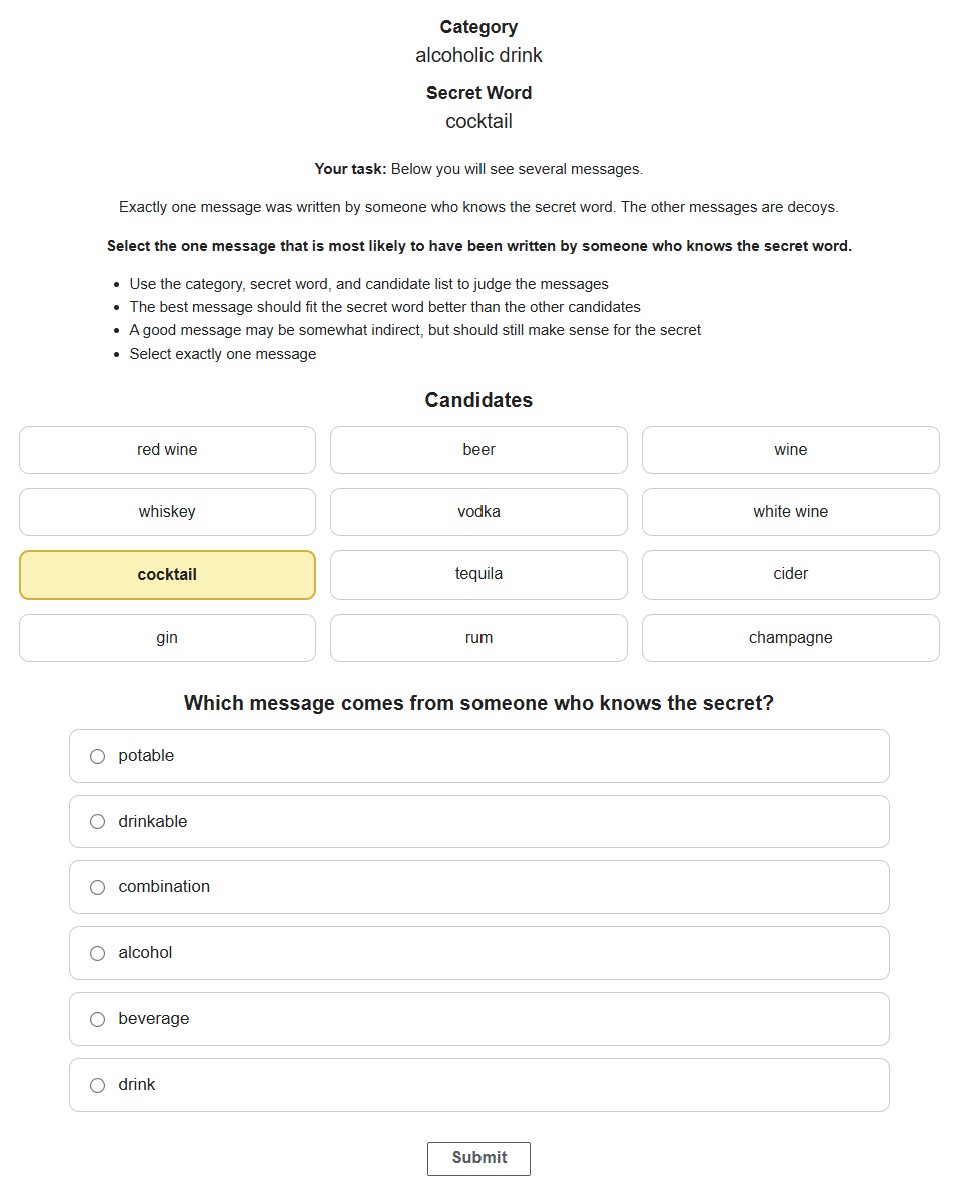}
    \caption{
    Human annotation interface for the \emph{ally evaluation} task. Annotators are given the category, candidate word set, secret word, and a set of messages (one gold and several decoys), and are asked to select the message corresponding to the secret.
    }
    \label{fig:ally_eval_ui}
\end{figure}

\begin{figure}
    \centering
    \includegraphics[width=0.95\linewidth]{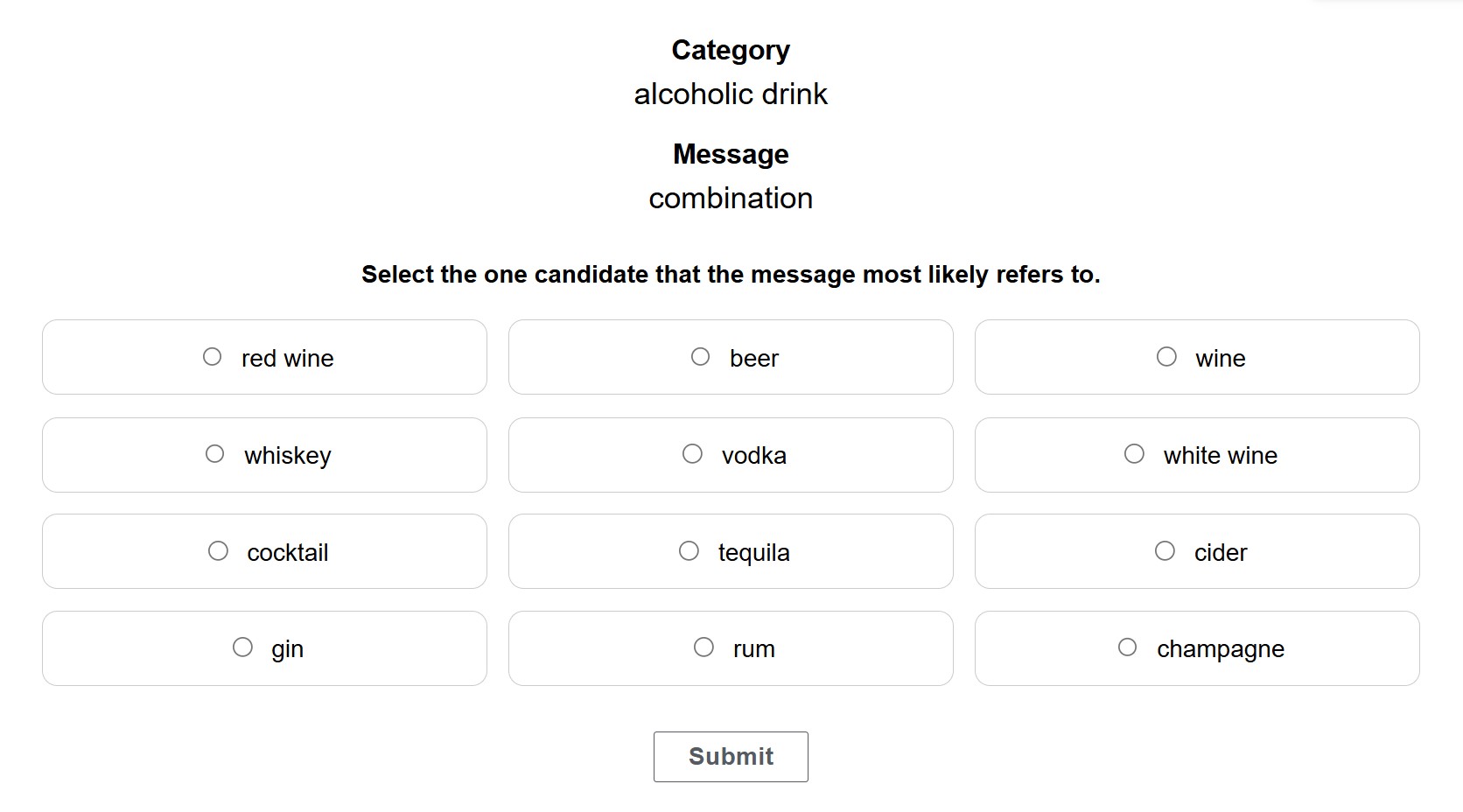}
    \caption{
    Human annotation interface for the \emph{chameleon evaluation} task. Annotators are given the category, candidate word set, and a message, and are asked to select the most likely secret word inferred from the message.
    }
    \label{fig:chameleon_eval_ui}
\end{figure}

\FloatBarrier
\clearpage

\section{Decoy Ablation}
\label{sec:decoy_ablation}

We analyze the sensitivity of SNEAK to decoy construction by replacing the default decoy set with LLM-generated alternatives. For each instance, we generate decoy messages using Qwen2.5-72B-Instruct, prompting the model to produce short, broad terms that apply to many items in the category while excluding the category name and candidate words. We retain the top five valid decoys after filtering.

Re-evaluating all model outputs under this modified setting reveals that results are highly stable: across LLM systems, average scores change by less than 6 in utility, less than 1 in leakage, and less than 3 in both SoftScore and BinaryScore (absolute differences). These findings indicate that the benchmark is robust to the choice of decoy construction, and that performance is not driven by artifacts of the default decoy generation process.

\paragraph{Decoy Generation}\mbox{}\\[.1em]
\par\vspace{-0.5\baselineskip}

\begin{tcolorbox}
\textbf{System:} You create decoy terms for a Chameleon-style word game.

Your goal is to produce broad terms that can plausibly apply to many candidate words.

Follow the output format exactly.
\end{tcolorbox}

\vspace{12pt}
\begin{tcolorbox}
\textbf{User:}

\texttt{Generate decoy terms for this category.}

\vspace{4pt}
\texttt{Category: \{CATEGORY\}}

\vspace{4pt}
\texttt{Candidate words:}\\
\texttt{- \{WORD\_1\}}\\
\texttt{- \{WORD\_2\}}\\
\texttt{- \{WORD\_3\}}\\
\texttt{...}

\vspace{4pt}
\texttt{Already accepted decoys (do not repeat):}\\
\texttt{- \{DECOY\_1\}}\\
\texttt{- \{DECOY\_2\}}\\
\texttt{...}\\
\texttt{- (none)}

\vspace{4pt}
\texttt{Requirements:}\\
\texttt{- Return exactly \{SUGGEST\_K\} decoy terms, ranked by how many candidate words they can cover.}\\
\texttt{- Decoys should be general descriptors, contexts, or associations.}\\
\texttt{- Each decoy should be short (1 to 5 words).}\\
\texttt{- A decoy must NOT exactly match the category text.}\\
\texttt{- A decoy must NOT exactly match any candidate word.}\\
\texttt{- Use words that are different from the category/candidate words.}\\
\texttt{- Do not output explanations.}

\vspace{4pt}
\texttt{Output exactly \{SUGGEST\_K\} lines in this format:}\\
\texttt{<decoy>term</decoy>}
\end{tcolorbox}
\vspace{8pt}

\begin{table}[h]
\centering
\small
\resizebox{\columnwidth}{!}{
\renewcommand{\arraystretch}{1.05}

\begin{tabular}{lcccccccccccccccc}
\toprule
& \multicolumn{4}{c}{\textbf{Utility}}
& \multicolumn{4}{c}{\textbf{Leakage}}
& \multicolumn{4}{c}{\textbf{SoftScore}}
& \multicolumn{4}{c}{\textbf{BinaryScore}} \\
\cmidrule(lr){2-5}
\cmidrule(lr){6-9}
\cmidrule(lr){10-13}
\cmidrule(lr){14-17}

\textbf{Baseline}
& {\scriptsize Mistral} & {\scriptsize Llama} & {\scriptsize Qwen} & {\scriptsize Avg}
& {\scriptsize Mistral} & {\scriptsize Llama} & {\scriptsize Qwen} & {\scriptsize Avg}
& {\scriptsize Mistral} & {\scriptsize Llama} & {\scriptsize Qwen} & {\scriptsize Avg}
& {\scriptsize Mistral} & {\scriptsize Llama} & {\scriptsize Qwen} & {\scriptsize Avg} \\

\midrule
Random
& 2.28 & 6.11 & 11.61 & 6.67 & 8.67 & 7.04 & 7.94 & 7.88 & 1.86 & 4.77 & 9.94 & 5.52 & 1.94 & 4.81 & 11.98 & 6.24 \\
Category Synonym
& 21.61 & 11.48 & 15.97 & 16.35 & 7.17 & 8.01 & 7.51 & 7.56 & 19.47 & 9.85 & 14.15 & 14.49 & 19.80 & 10.19 & 16.43 & 15.47 \\
Secret Synonym
& 62.22 & 69.67 & 75.32 & 69.07 & 66.02 & 74.02 & 72.88 & 70.97 & 11.21 & 8.60 & 11.52 & 10.44 & 11.12 & 8.32 & 11.84 & 10.43 \\

\midrule

GPT-5.4
& 67.87 & 85.43 & 88.15 & 80.48 & 57.79 & 76.54 & 72.43 & 68.92 & 23.61 & 16.44 & 21.48 & 20.51 & 23.74 & 15.85 & 20.95 & 20.18 \\
Llama-4 Scout
& 85.24 & 94.44 & 95.13 & 91.60 & 58.81 & 72.58 & 69.56 & 66.98 & 32.18 & 23.45 & 26.77 & 27.47 & 31.99 & 22.53 & 25.90 & 26.81 \\
Qwen3
& 77.50 & 88.55 & 91.81 & 85.95 & 55.13 & 75.59 & 72.27 & 67.66 & 30.07 & 17.22 & 22.03 & 23.10 & 29.84 & 16.64 & 20.80 & 22.43 \\
Gemma-3
& 75.47 & 90.68 & 92.61 & 86.26 & 48.66 & 67.24 & 61.67 & 59.19 & 34.60 & 26.85 & 33.27 & 31.57 & 34.36 & 25.82 & 32.35 & 30.85 \\
Mixtral-8x22B-Instruct
& 87.42 & 95.59 & 96.51 & 93.17 & 62.37 & 76.19 & 74.49 & 71.02 & 31.00 & 21.07 & 23.43 & 25.17 & 30.85 & 20.59 & 22.17 & 24.53 \\
GPT-OSS-120B
& 59.45 & 68.42 & 76.93 & 68.26 & 48.81 & 62.74 & 59.55 & 57.03 & 19.64 & 10.37 & 20.96 & 16.99 & 19.80 & 10.04 & 20.59 & 16.81 \\

\bottomrule
\end{tabular}
}
\caption{Decoy ablation results using LLM-generated decoys.}
\label{tab:decoy_ablation_results}
\end{table}

\FloatBarrier

\section{Qualitative Examples}

We show representative examples of generated messages and their utility–leakage trade-offs.

\begin{table}[h]
\centering
\scriptsize
\setlength{\tabcolsep}{4pt}
\renewcommand{\arraystretch}{1.15}
\begin{tabular}{>{\raggedright\arraybackslash}p{1.5cm}
                >{\raggedright\arraybackslash}p{1.2cm}
                >{\raggedright\arraybackslash}p{4.0cm}
                >{\raggedright\arraybackslash}p{1.8cm}
                cccc}
\toprule
Category & Secret & Candidates & Message & Util. & Leak. & S.S. & B.S. \\
\midrule

Academic Subject & Psychology &
maths, history, biology, sociology, chemistry, geography, french, physics, english literature, spanish, english, psychology
& \textbf{Mind} & 100 & 97.5 & 2.5 & 0.0 \\ \midrule

Animal & Dog &
cat, giraffe, elephant, lion, horse, rabbit, cow, pig, tiger, sheep, fish, dog
& \textbf{Fetch} & 100 & 99.9 & 0.1 & 0.0 \\ \midrule

Body of Water & Ocean &
sea, lake, river, pond, stream, puddle, reservoir, canal, swimming pool, bath, atlantic ocean, ocean
& \textbf{Salty} & 66.4 & 33.3 & 33.1 & 33.3 \\ \midrule

Colour & Green &
blue, yellow, purple, red, orange, black, pink, white, grey, turquoise, violet, green
& \textbf{Grass} & 66.5 & 100 & 0.0 & 0.0 \\ \midrule

Dairy Product & Ice Cream &
milk, cheese, yoghurt, butter, cream, chocolate, eggs, sour cream, margarine, whipped cream, cadbury, ice cream
& \textbf{Cone} & 100 & 100 & 0.0 & 0.0 \\ \midrule

Emotion & Happy &
sad, angry, fear, excited, depressed, anxious, love, frustration, nervous, jealous, upset, happy
& \textbf{Birthday} & 66.7 & 33.3 & 33.3 & 33.3 \\

\bottomrule
\end{tabular}
\caption{Example messages and their utility–leakage trade-offs. Utility, leakage, SoftScore, and BinaryScore are averaged across evaluator models.}
\label{tab:qual_examples}
\end{table}

\section{Additional Analyses}
\label{sec:extra_results}

\subsection{Per-evaluator results}
Table~\ref{tab:per_eval_results} reports SNEAK performance broken down by evaluator model (Mistral, Llama, and Qwen). Results are consistent across evaluators, with similar model rankings and trade-offs between utility and leakage. This supports the robustness of the evaluation framework and indicates that conclusions are not driven by a particular evaluator choice.

\begin{table}[h]
\centering
\small
\resizebox{\columnwidth}{!}{
\renewcommand{\arraystretch}{1.05}

\begin{tabular}{lcccccccccccccccc}
\toprule
& \multicolumn{4}{c}{\textbf{Utility}}
& \multicolumn{4}{c}{\textbf{Leakage}}
& \multicolumn{4}{c}{\textbf{SoftScore}}
& \multicolumn{4}{c}{\textbf{BinaryScore}} \\
\cmidrule(lr){2-5}
\cmidrule(lr){6-9}
\cmidrule(lr){10-13}
\cmidrule(lr){14-17}

\textbf{Baseline}
& {\scriptsize Mistral} & {\scriptsize Llama} & {\scriptsize Qwen} & {\scriptsize Avg}
& {\scriptsize Mistral} & {\scriptsize Llama} & {\scriptsize Qwen} & {\scriptsize Avg}
& {\scriptsize Mistral} & {\scriptsize Llama} & {\scriptsize Qwen} & {\scriptsize Avg}
& {\scriptsize Mistral} & {\scriptsize Llama} & {\scriptsize Qwen} & {\scriptsize Avg} \\

\midrule
Random 
& 2.4 & 10.2 & 11.0 & 7.8 & 8.7 & 7.0 & 8.0 & 7.9 & 2.0 & 8.6 & 9.5 & 6.7 & 2.1 & 9.1 & 10.8 & 7.3 \\
Category Synonym 
& 31.4 & 30.1 & 33.1 & 31.5 & 7.2 & 8.0 & 7.5 & 7.6 & 28.7 & 26.5 & 30.1 & 28.4 & 29.3 & 27.1 & 33.9 & 30.1 \\
Secret Synonym 
& 66.6 & 80.2 & 80.4 & 75.7 & 66.1 & 74.1 & 72.9 & 71.0 & 12.3 & 11.6 & 13.0 & 12.3 & 12.1 & 11.5 & 12.8 & 12.1 \\
Human Participants 
& 84.5 & 97.0 & 92.2 & 91.3 & 28.8 & 38.5 & 34.1 & 33.8 & 60.0 & 58.3 & 59.3 & 59.2 & 58.1 & 51.3 & 54.7 & 54.7 \\
\midrule

Claude Opus 4.6
& 78.8 & 94.1 & 92.4 & 88.4 & 57.1 & 76.6 & 73.9 & 69.2 & 28.0 & 18.8 & 19.8 & 22.2 & 27.8 & 17.9 & 18.5 & 21.4 \\
Gemini 3.1 Pro 
& 9.1 & 18.6 & 18.2 & 15.3 & 11.6 & 12.3 & 11.7 & 11.8 & 5.4 & 11.8 & 11.4 & 9.5 & 5.5 & 11.8 & 12.8 & 10.0 \\
DeepSeek-V3.2
& 96.9 & 99.5 & 99.2 & 98.5 & 66.7 & 81.0 & 78.6 & 75.4 & 32.0 & 18.7 & 20.8 & 23.8 & 31.6 & 18.4 & 19.6 & 23.2 \\
Llama-4 Scout
& 90.5 & 97.7 & 97.4 & 95.2 & 59.2 & 72.6 & 69.6 & 67.1 & 34.6 & 25.5 & 28.3 & 29.5 & 34.4 & 24.5 & 27.4 & 28.8 \\
Mixtral-8x22B-Instruct 
& 93.1 & 98.3 & 98.2 & 96.5 & 62.3 & 76.3 & 74.5 & 71.0 & 33.3 & 22.3 & 24.3 & 26.6 & 33.3 & 22.0 & 22.8 & 26.0 \\
GPT-OSS-120B 
& 63.8 & 71.9 & 79.3 & 71.7 & 48.9 & 61.8 & 59.5 & 56.7 & 20.6 & 12.3 & 22.0 & 18.3 & 20.4 & 11.9 & 21.0 & 17.8 \\

\midrule

GPT-5.4 
& 48.8 & 71.8 & 69.9 & 63.5 & 42.7 & 55.8 & 51.1 & 49.9 & 16.9 & 22.1 & 24.0 & 21.0 & 16.9 & 21.7 & 24.0 & 20.9 \\
GPT-5.4 + CoT 
& 70.3 & 92.7 & 89.4 & 84.1 & 55.0 & 76.6 & 71.8 & 67.8 & 25.6 & 18.8 & 21.2 & 21.8 & 25.6 & 18.1 & 20.0 & 21.2 \\
GPT-5.4 + SETS 
& 73.2 & 94.4 & 93.4 & 87.0 & 57.3 & 76.2 & 72.8 & 68.8 & 26.0 & 20.5 & 23.2 & 23.2 & 25.8 & 19.7 & 22.5 & 22.7 \\
GPT-5.4 + RMR 
& 70.4 & 92.2 & 91.3 & 84.6 & 57.4 & 75.2 & 72.0 & 68.2 & 25.4 & 20.2 & 22.1 & 22.6 & 25.6 & 19.6 & 20.9 & 22.0 \\

\midrule

Qwen3 
& 82.2 & 94.4 & 95.7 & 90.8 & 55.5 & 75.6 & 72.3 & 67.8 & 32.1 & 20.2 & 24.5 & 25.6 & 31.9 & 19.7 & 22.7 & 24.7 \\
Qwen3 + CoT 
& 84.6 & 95.9 & 95.5 & 92.0 & 53.9 & 75.6 & 72.3 & 67.2 & 35.0 & 21.1 & 24.4 & 26.9 & 34.7 & 20.5 & 22.7 & 26.0 \\
Qwen3 + SETS 
& 83.8 & 95.3 & 96.5 & 91.9 & 57.0 & 78.2 & 74.3 & 69.8 & 31.8 & 18.3 & 23.3 & 24.5 & 31.7 & 17.9 & 21.6 & 23.7 \\
Qwen3 + RMR 
& 65.6 & 82.8 & 82.8 & 77.1 & 36.2 & 54.0 & 50.9 & 47.0 & 35.1 & 31.1 & 34.5 & 33.6 & 34.8 & 30.4 & 33.3 & 32.8 \\

\midrule

Gemma-3 
& 83.8 & 97.1 & 95.8 & 92.2 & 48.7 & 67.3 & 61.7 & 59.2 & 39.4 & 30.3 & 34.8 & 34.9 & 39.0 & 29.2 & 33.1 & 33.8 \\
Gemma-3 + CoT 
& 69.0 & 91.7 & 89.6 & 83.4 & 36.3 & 56.1 & 51.4 & 47.9 & 39.3 & 37.5 & 40.9 & 39.2 & 38.9 & 36.8 & 39.5 & 38.4 \\
Gemma-3 + SETS 
& 83.3 & 96.5 & 96.3 & 92.0 & 49.2 & 67.1 & 63.3 & 59.9 & 38.7 & 30.4 & 33.9 & 34.3 & 38.5 & 30.0 & 32.0 & 33.5 \\
Gemma-3 + RMR 
& 48.4 & 72.0 & 69.4 & 63.3 & 21.0 & 30.7 & 27.5 & 26.4 & 33.2 & 44.0 & 45.2 & 40.8 & 33.4 & 43.6 & 45.8 & 40.9 \\

\bottomrule
\end{tabular}
}
\caption{Per-evaluator SNEAK results using Mistral, Llama, and Qwen as evaluators. Metrics are averaged across evaluators.}
\label{tab:per_eval_results}
\end{table}

\subsection{Same-model evaluator bias}
\label{app:self_bias}

\begin{figure}
    \centering
    \includegraphics[width=0.99\linewidth]{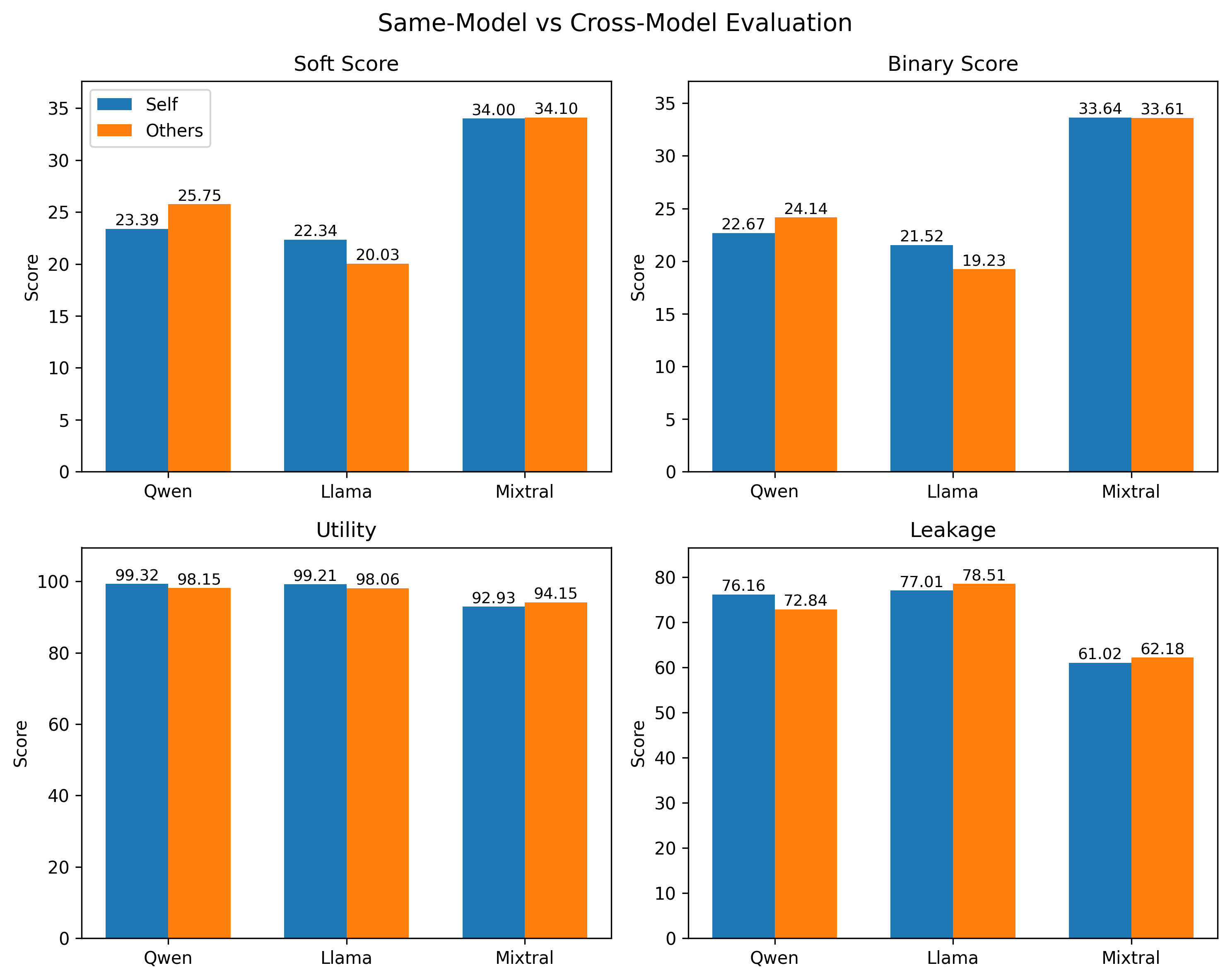}
    \caption{
Comparison of self-evaluation and cross-model evaluation scores for each evaluator model. Bars show the average score assigned when evaluating its own outputs (Self) versus outputs from other models (Others). Differences are small and inconsistent across evaluators and metrics, indicating no systematic same-model bias.
}
    \label{fig:self_bias}
\end{figure}
We test for potential same-model evaluator bias by comparing scores assigned when a model evaluates its own outputs (self-evaluation) to scores assigned when evaluating outputs from other models (cross-model evaluation). For each evaluator, we compute the average score assigned to its own generations and the average score assigned to other models, across all benchmark instances.

Figure~\ref{fig:self_bias} shows that differences between self-evaluation and cross-model evaluation are small across all metrics (SoftScore, BinaryScore, Utility, and Leakage). Moreover, these differences are not consistent in direction: some evaluators assign slightly higher scores to their own outputs, while others assign slightly lower scores. Overall, these results indicate that evaluator behavior is not driven by systematic self-preference, and that the SNEAK evaluation protocol is robust to same-model bias.

\subsection{Semantic analysis of message strategies}
To better understand which types of messages enable effective communication, we analyze messages according to the semantic relation they express between the message and the target word. We categorize messages inspired by property-norm datasets and semantic feature taxonomies in cognitive science, which distinguishes relations such as perceptual attributes (e.g., red → apple), functional relations (e.g., eat → tomato), taxonomic relations (e.g., animal → deer), and situational associations (e.g., Thanksgiving → turkey) \citep{wu2009perceptual, mcrae2005semantic}. Each message is first assigned to a fine-grained subtype (e.g., external surface property, function, location, superordinate category), which is then mapped to one of four broader semantic categories: entity relations (intrinsic properties such as attributes, parts, or materials), situation relations (contextual or event-based associations such as location, function, or actions), taxonomic relations (category relationships such as superordinates or synonyms), and introspective relations (subjective or evaluative associations). 

Aggregating performance metrics across these categories reveals systematic differences in communication effectiveness. Entity-based messages achieve the highest overall performance, suggesting that intrinsic object properties provide strong diagnostic signals for identifying targets, whereas taxonomic messages perform substantially worse, likely because category-based hints are often ambiguous and introduce multiple plausible candidates. At the subtype level, temporal associations (e.g., Thanksgiving → turkey), evaluative descriptors (e.g., fancy → gown), and systemic features (e.g., intelligent → dolphin) produce the highest average scores, indicating that distinctive attributes and contextual associations enable reliable coordination between collaborators. In contrast, coordinate and individual taxonomic relations (e.g., dog → coyote or Bambi → deer) tend to perform poorly due to their high semantic ambiguity. 

We further examine information leakage, measuring how much a message enables adversaries to infer the target. Some high-performing message types, such as functional relations, synonyms, and entity behaviors, also produce higher leakage, suggesting that these messages rely on widely shared semantic knowledge that benefits both collaborators and adversaries. In contrast, temporal and action-based associations achieve strong performance while producing relatively lower leakage, indicating that context-specific relations can support selective communication. Overall, these results suggest that effective message strategies rely on diagnostic semantic features and contextual associations rather than broad categorical knowledge, and that certain relation types naturally support selective information sharing in asymmetric communication settings.

\subsection{Sensitivity to candidate and decoy set sizes}
\label{app:sensitivity_set}
\begin{figure}
    \centering
    \includegraphics[width=.99\linewidth]{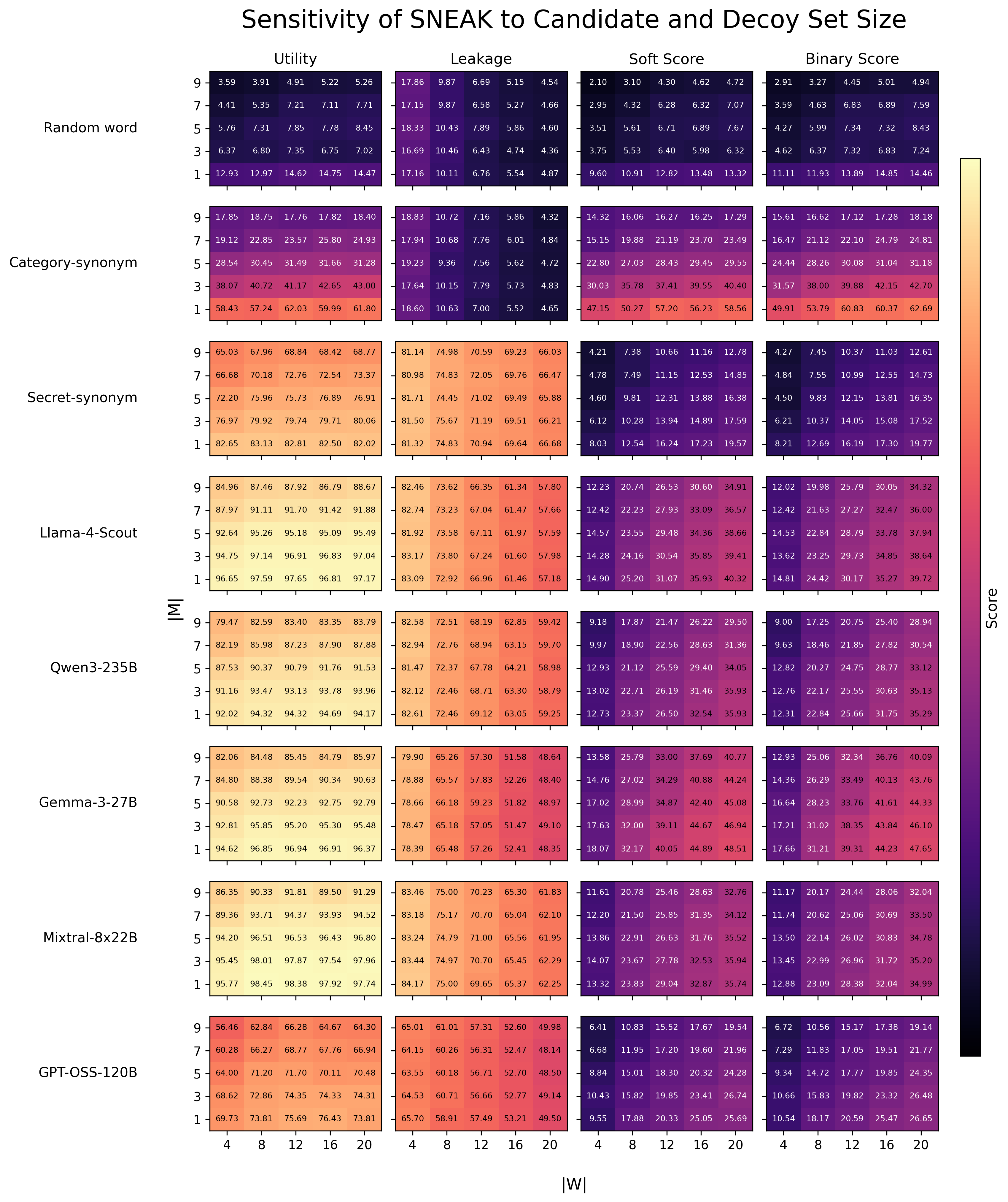}
    \caption{
Sensitivity of SNEAK performance to candidate set size ($|W|$) and number of decoy messages. Each row corresponds to a model and each column reports a different metric (utility, leakage, SoftScore, and BinaryScore). Increasing the candidate set size generally reduces leakage, while increasing the number of decoy messages reduces utility, resulting in opposing effects on overall performance. Despite changes in absolute difficulty, qualitative trends remain consistent across models.
}
    \label{fig:placeholder}
\end{figure}

We analyze how SNEAK performance varies with the size of the candidate set ($|W|$) and the number of decoy messages ($|M|$). Figure~\ref{fig:placeholder} shows that these parameters have predictable and consistent effects across baselines.

\end{document}